\documentclass[balance,upint,subscriptcorrection,varvw,mathalfa=cal=boondoxo,pdf-a,colorlinks]{_asmeconf/asmeconf}

\definecolor{darkgreen}{rgb}{0.0, 0.5, 0.0}
\definecolor{blue}{rgb}{0.0, 0.47, 0.75}
\definecolor{dartmouthgreen}{rgb}{0.05, 0.5, 0.06}
\definecolor{drab}{rgb}{0.59, 0.44, 0.09}
\definecolor{navyblue}{rgb}{0.0, 0.0, 0.5}

\definecolor{darkgreen}{rgb}{0.0, 0.5, 0.0}
\definecolor{blue}{rgb}{0.0, 0.47, 0.75}
\definecolor{dartmouthgreen}{rgb}{0.05, 0.5, 0.06}
\definecolor{drab}{rgb}{0.59, 0.44, 0.09}
\definecolor{navyblue}{rgb}{0.0, 0.0, 0.5}

\newcommand{\KL}{\mathbb{KL}}

\newcommand{\vc}{\mathbf{c}}

\newcommand{\vg}{\mathbf{g}}

\newcommand{\vx}{\mathbf{x}}

\newcommand{\vz}{\mathbf{z}}

\newcommand{\vtheta}{\mathbf{\theta}}

\newcommand{\cN}{\mathcal{N}}

\newcommand{\bE}{\mathbb{E}}

\newcommand{\bR}{\mathbb{R}}

\usepackage[dvipsnames]{xcolor}
\usepackage{pifont}%
\newcommand{\cmark}{ {\color{dartmouthgreen} \ding{51}} }%
\newcommand{\xmark}{ {\color{red} \ding{55}} }%

\begin{document}

\title{Diffusing the Optimal Topology: A Generative Optimization Approach}

\SetAuthors{%
Giorgio Giannone\affil{1}\affil{2}\CorrespondingAuthor{ggiorgio@mit.edu, faez@mit.edu}, 
	Faez Ahmed\affil{1} %
 }

\SetAffiliation{1}{Massachusetts Institute of Technology, Cambridge, MA }
\SetAffiliation{2}{Technical University of Denmark, Lyngby, DK}

\maketitle

\begin{abstract}
Topology Optimization seeks to find the best design that satisfies a set of constraints while maximizing system performance. Traditional iterative optimization methods like SIMP can be computationally expensive and get stuck in local minima, limiting their applicability to complex or large-scale problems. Learning-based approaches have been developed to accelerate the topology optimization process, but these methods can generate designs with floating material and low performance when challenged with out-of-distribution constraint configurations. Recently, deep generative models, such as Generative Adversarial Networks and Diffusion Models, conditioned on constraints and physics fields have shown promise, but they require extensive pre-processing and surrogate models for improving performance. To address these issues, we propose a Generative Optimization method that integrates classic optimization like SIMP as a refining mechanism for the topology generated by a deep generative model. We also remove the need for conditioning on physical fields using a computationally inexpensive approximation inspired by classic ODE solutions and reduce the number of steps needed to generate a feasible and performant topology. Our method allows us to efficiently generate good topologies and explicitly guide them to regions with high manufacturability and high performance, without the need for external auxiliary models or additional labeled data. We believe that our method can lead to significant advancements in the design and optimization of structures in engineering applications, and can be applied to a broader spectrum of performance-aware engineering design problems.
\end{abstract}

\section{Introduction}
\begin{figure}[ht]
    \centering
    \includegraphics[width=.95\linewidth]{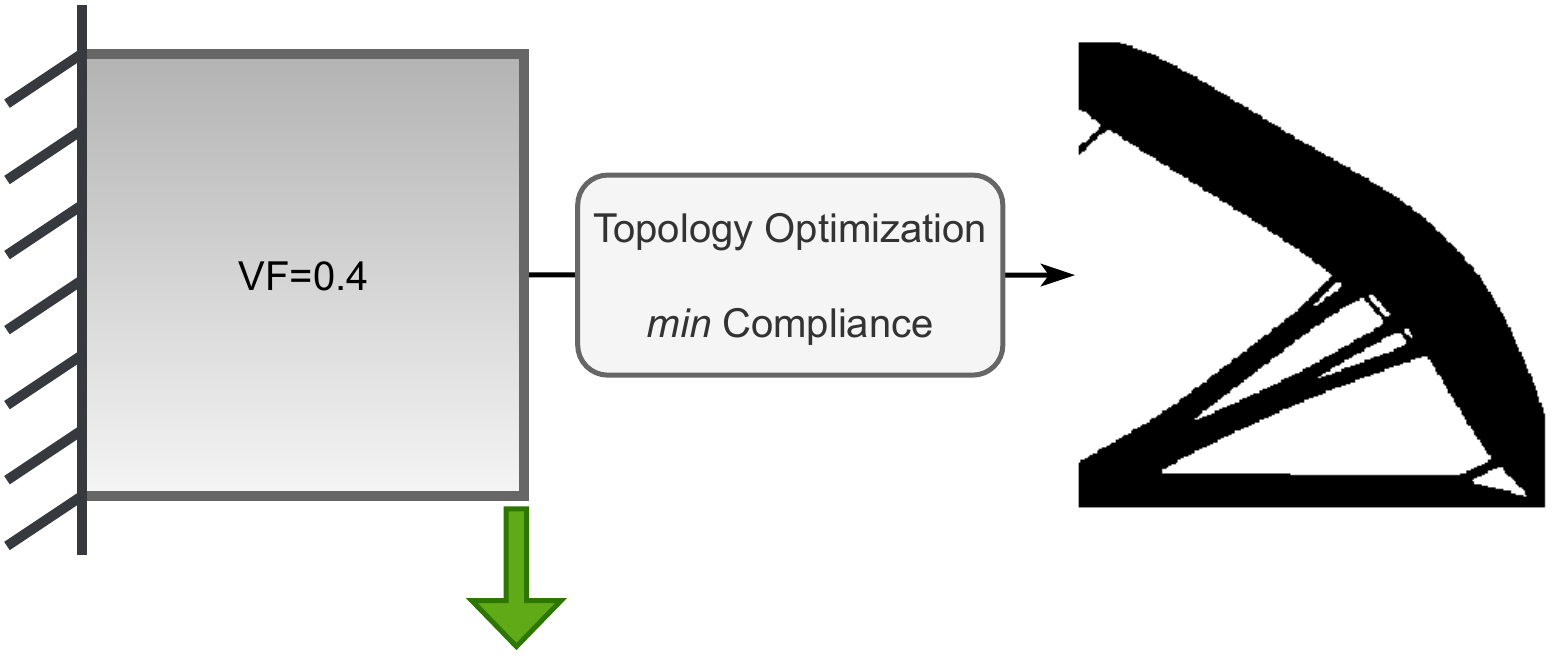} 
    \caption{Minimum Compliance Topology Optimization problem. 
    Given a set of constraints (loads, boundary conditions, volume fraction), find the optimal topology minimizing functional performance (structural compliance).}
    \label{fig:intro_to}
\end{figure}

Topology optimization~(TO \citep{bendsoe1988generating, sigmund2013topology}) is an essential engineering tool that enables finding the optimal material distribution for meeting performance objectives while satisfying constraints. Constraints can take various forms, such as loads, boundary conditions, and volume fraction. Topology optimization can lead to significant improvements in the efficiency, safety, reliability, and durability of structures across various fields, including aerospace, automotive, civil, and mechanical engineering. The demand for improvements in topology optimization techniques is ever-growing to enhance design engineering for a new age of engineering discovery.
Topology optimization uses Finite Element Analysis (FEA), with gradient-based~\cite{bendsoe1988generating} and gradient-free methods~\cite{ahmed2013constructive} being two primary approaches. The Solid Isotropic Material with Penalization (SIMP) method~\cite{bendsoe1989optimal, rozvany1992generalized} is one of the most widely used gradient-based optimization algorithms. SIMP utilizes iterative optimization methods over continuous, density-based representation to determine the optimal material distribution. This class of methods iteratively adjusts the material density in a given domain to check whether the design meets the desired constraints.

Iterative optimization-based topology optimization (TO) methods have great potential benefits but face challenges in practical applications, particularly for large-scale problems due to their computational complexity. Iterative algorithms, like SIMP, are computationally expensive and prone to getting trapped in local optima, limiting the search for a global optimum. To address these challenges, there has been a recent surge of interest in learning-based methods for topology optimization, which use machine learning (ML) algorithms, such as deep generative models, to speed up the optimization process and generate more diverse structural topologies. By leveraging large datasets of existing designs, ML models can quickly identify patterns and generate new designs that meet specified constraints. Compared to traditional optimization methods, ML methods can handle high-dimensional data, explore a broader range of design solutions, and provide more diverse results that may not have been considered previously.
Companies like SolidWorks and start-ups like nTopology are integrating deep generative topology optimization in their products, adding flexibility and creativity for the designers and speeding up the candidate generation phase.

\paragraph{Deep Generative Models for TO.}
Deep Generative models (DGMs~\cite{bond2021deep}), have shown an impressive ability to model high-dimensional and complex data modalities, such as images or text, with high fidelity and diversity. This capability opens the door for significant improvements in creativity and productivity in various fields in the coming years.
Recent advancements in large vision models~\cite{rombach2021high}) and language models~\cite{brown2020language}) have greatly increased our capacity to process unstructured data, leading to the development of multimodal generation techniques~\cite{ramesh2021zero}. These models hold great promise for enhancing engineering design~\cite{regenwetter2022deep, song2023multi}.
Building on these successes, DGMs have also been applied in the engineering field to problems with constraints, with a focus on improving the design process. However, these applications have primarily focused on metrics like reconstruction quality and have not fully addressed the fulfillment of engineering requirements.

DGMs for topology optimization pose challenges in terms of both performance and manufacturability, which TopologyGAN~\citep{nie2021topologygan}, addresses by conditioning on physics-based information. By introducing physical fields such as the Von Mises stress, strain energy density, and displacement fields, TopologyGAN improves the adherence of the generated samples to the underlying engineering problem. However, computing the physical fields requires solving a computationally expensive Finite Element Analysis (FEA) routine for each configuration of load and boundary condition considered, even though sampling new topologies with the base model (a conditional GAN) is fast.
This approach has been successful but still optimizes for a pixel-wise, reconstruction-based loss that does not account for engineering requirements such as high performance and feasibility.
Despite conditioning on physical fields, many designs generated by these models still suffer from floating materials that impede manufacturability, as well as limited diversity and generalizability out-of-distribution.

TopoDiff~\cite{maze2022topodiff} proposes a conditional diffusion model, conditioning on fields similarly to~\cite{nie2021topologygan}, and an additional guidance mechanism to encourage the generative process to sample in regions with high manufacturability and high performance (low compliance in this specific case). This work solves some of the issues connected with performance and feasibility, at the cost of slow sampling, still relying on computationally expensive physical fields, and introducing surrogate models to account for performance and manufacturability.
During inference, the model must compute the strain and force fields for each constraint configuration using FEA and use this information to condition the model, enabling it to generate optimized topologies that meet the specified requirements. This conditioning step is crucial to ensure that the model performs well and generates output that is both feasible and optimized but, at the same time, expensive and not easy to scale to higher dimensionality, complex structures, and 3D domains.

To summarize, GAN-based methods such as those proposed in \cite{nie2021topologygan, Behzadi_Ilies2021} can generate a large number of topologies efficiently but may produce un-manufacturable topologies that violate soft constraints such as volume fraction errors and lead to higher compliance structures. On the other hand, diffusion-based approaches like TopoDiff~\cite{maze2022topodiff} generate samples that satisfy constraints more accurately but are computationally expensive due to iterative sampling, reliance on physical fields, and surrogate models with auxiliary labeled data.

\paragraph{}
To overcome the limitations of existing approaches, we propose a novel method that addresses the issues of slow sampling, reliance on physical fields, and the need for additional surrogate models. Our approach involves reducing the number of steps required for sampling, approximating physical fields using a computationally inexpensive kernel based on classic ODE solutions, and integrating optimization methods like SIMP for refining the generated topology. By explicitly guiding the generated topology to regions with high manufacturability and low compliance with only a few optimization steps, we can efficiently sample good topologies without the need for external auxiliary models, FEM solvers for pre-processing, or additional labeled data.

\paragraph{Contribution.} Our contributions are the following:

\begin{itemize}
    \item We decrease the computational cost of generative topology optimization while maintaining high performance. To achieve this, we explore more efficient sampling methods for TopoDiff reducing the number of steps required for generation by an order of magnitude while ensuring minimal loss in performance across all models.
    \item We explore alternative conditioning techniques that eliminate the need to compute force and energy strain fields, which can be a major bottleneck in the optimization process. As a result, we have reduced the inference time for generation by 53.93\% compared to the baselines.
    \item We propose a generative optimization method that integrates a conditional diffusion model with traditional topology optimization algorithms. The fast conditional diffusion model predicts a preliminary topology, which is then refined using traditional topology optimization-based methods in just a few steps (5-10 iterations). This approach improves manufacturability and increases performance by 23.81\% and 25.64\%, respectively, for in- and out-of-distribution constraints.
\end{itemize}

\section{Related Work}
\begin{figure}[ht]
    \centering
    \includegraphics[width=.95\linewidth]{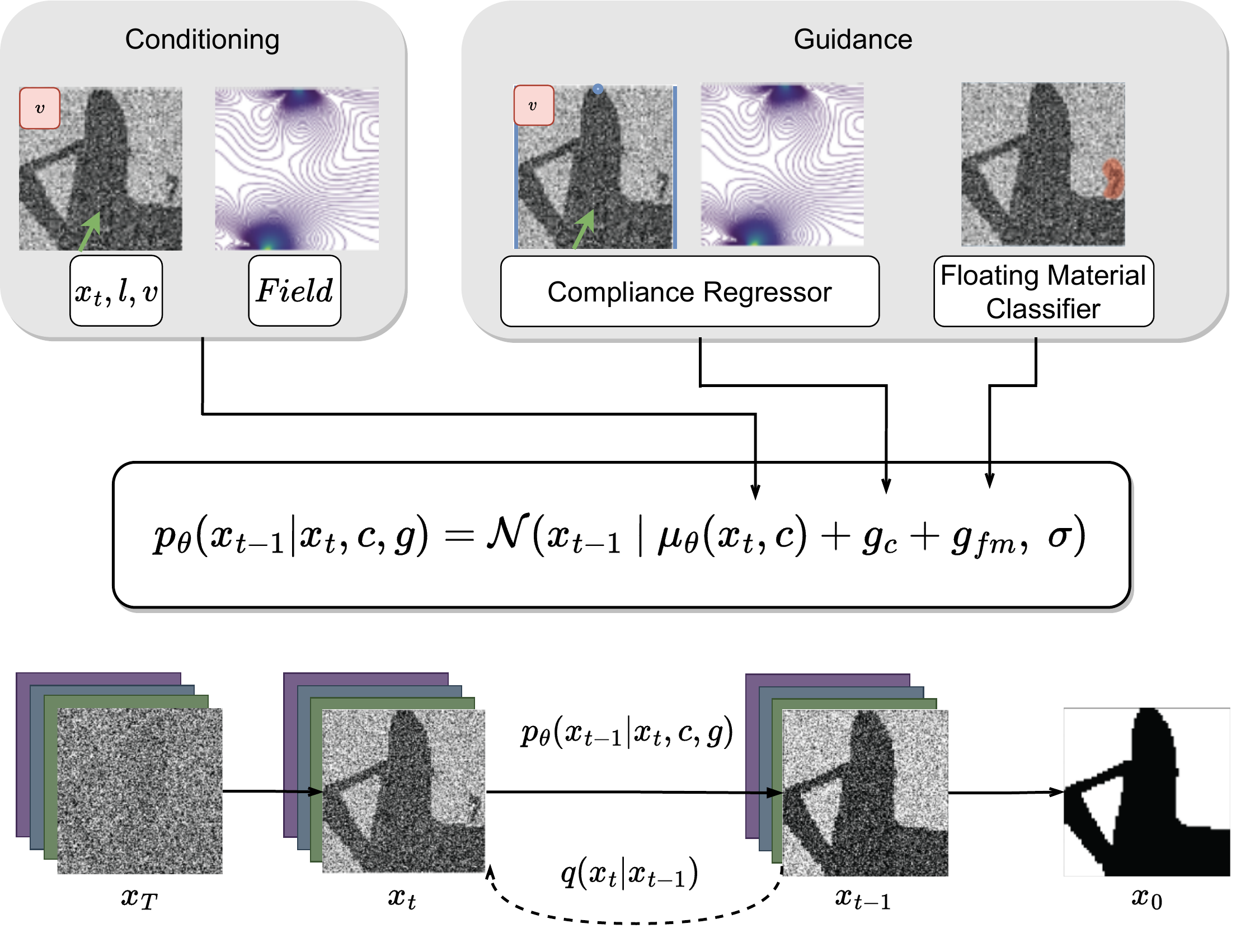}  
    \caption{TopoDiff-GUIDED. TopoDiff is a conditional diffusion model guided by a classifier and a regression score. The conditioning mechanism $c$ involves loads $l$, volume fraction $v$, and stress and energy fields $f$. i.e. $c=h(l, v, f)$. The guidance mechanism involves a classifier for the presence or absence of floating material and a regressor to minimize compliance error between generated and optimized topologies.}
    \label{fig:topodiff-intro}
\end{figure}
\vspace{5pt}
\paragraph{Topology Optimization.}
Engineering design is the process of creating solutions to technical problems under engineering requirements~\cite{shigley1985mechanical}. The goal is to create designs that are highly performant given the required constraints.
Topology Optimization (TO~\citep{bendsoe1988generating}) is a branch of engineering design and is a critical component of the design process in many industries, including aerospace, automotive, manufacturing, and software development.
From the inception of the homogenization method for TO, a number of different approaches have been proposed, including density-based~\citep{bendsoe1989optimal, rozvany1992generalized}, level-set~\citep{allaire2002level}, derivative-based~\citep{sokolowski1999topological}, evolutionary~\citep{xie1997basic}, and others~\cite{bourdin2003design}. 
The density-based methods are widely used and use a nodal-based representation where the level-set leverages shapes derivative to obtain the optimal topology.
Topology Optimization has evolved as a computationally intensive discipline, with the availability of efficient open-source code~\cite{hunter2017topy, liu2014efficient}. See~\cite{liu2014efficient} for more on this topic and ~\cite{sigmund2013topology} for a comprehensive review of the Topology Optimization field.

\paragraph{Deep Learning for Topology Optimization.}
Following the success of Deep Learning (DL) in vision, a surging interest arose recently for transferring these methods to the engineering field.
In particular, DL methods have been employed for direct-design~\cite{Abueiddaetal2020}, accelating the optimization process~\cite{Bangaetal2018}, optimizing the shape~\cite{Hertleinetal2021}, super-resolusion~\cite{Elingaard2021, Napieretal2020}, and 3D topologies~\cite{kench2021generating}. 
Among these methods, Deep Generative Models (DGMs) are especially appealing to improve design diversity in engineering design\cite{JiangChenFan2021_nano2, RawatShen2018}.
Additionally, DGMs have been used for Topology Optimization problems conditioning on constraints (loads, boundary conditions, volume fraction for the structural case), directly generating topologies~\cite{rawat2019application, sharpe2019topology} training dataset of optimized topologies, leveraging superresolution methods to improve fidelity~\cite{yu2019deep}, using filtering and iterative design approaches~\cite{berthelot2017began} to improve quality and diversity. Methods for 3D topologies have also been proposed~\citep{Behzadi_Ilies2021}.
Recently, GAN-based approaches conditioning on constraints and physical information have had success in modeling the TO problem~\cite{nie2021topologygan}.
For a comprehensive review and critique of the field, see~\cite{woldseth2022use}.

\paragraph{Conditional Diffusion Models.}
Improving sampling speed for diffusion models is an active research topic~\cite{kong2021fast}. Recently, distillation has been used to greatly reduce sampling steps~\cite{meng2022distillation}.
Methods to condition DDPM have been proposed, conditioning at inference time~\cite{choi2021ilvr}, learning a class-conditional score~\cite{song2020score}, explicitly conditioning on class information~\cite{nichol2021improved}, set-based features~\cite{giannone2022few}, and physical properties~\cite{xie2021crystal}.
Text-to-image diffusion models~\cite{ramesh2022hierarchical} have been recently proposed for a guided generation.
Recently, TopoDiff~\citep{maze2022topodiff} has shown that conditional diffusion models are effective for generating topologies that fulfill the constraints, and have high manufacturability and high performance. TopoDiff relies on physics information and surrogate models to guide the sampling of novel topologies with good performance.

\section{Background}

\begin{figure*}[ht]
    \centering
    \includegraphics[width=.9\linewidth]{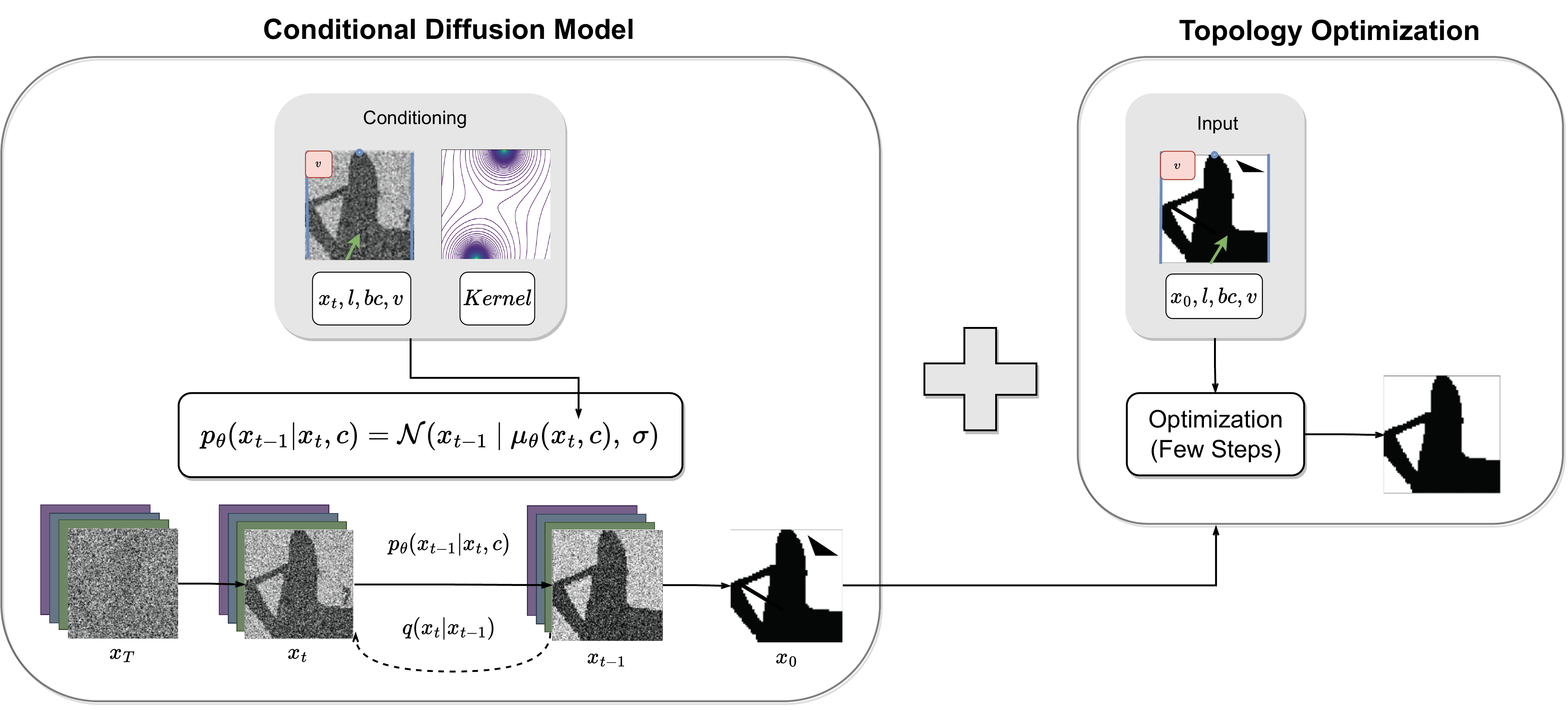} 
    \caption{TopoDiff-FF + SIMP. 
    The TopoDiff-FF pipeline is a conditional diffusion model where we condition on a cheap kernel relaxation instead of using expensive FEA to obtain the stress and energy field like in TopoDiff.
    After the generation step, we can improve the generated topology using few steps of SIMP (5/10 iterations) to remove floating material and explicitly optimize performance (minimize structural compliance).}
    \label{fig:topodiff_ff_simp}
\end{figure*}

Here we briefly introduce the Topology Optimization problem~\cite{bendsoe1988generating}, diffusion models~\citep{dickstein2015}, a class of deep generative models, conditioning and guidance mechanisms for diffusion models, and recent deep generative models for topology optimization~\cite{nie2021topologygan, maze2022topodiff}. 

\paragraph{The Topology Optimization Problem.}
Topology optimization (TO) is a powerful computational design approach used to determine the optimal configuration of a given structure, given a set of constraints. The goal of topology optimization is to identify the most efficient use of material while ensuring that the structure satisfies specific performance requirements.
One common approach to topology optimization is the SIMP (Solid Isotropic Material with Penalization) method~\citep{bendsoe1989optimal}. The SIMP method involves modeling the material properties using a density field, where the density represents the proportion of the material present in a given region. The optimization process involves adjusting the density field iteratively, subject to various constraints such as stress and boundary conditions.
\begin{table*}[ht]
    \centering
    \setlength{\tabcolsep}{4pt}
    \small
    \begin{tabular}{c|c c c c c c c c}
                    & Load & BC & Kernel Load & Kernel BC & Force Field & Energy Field & VF & Performance \\
        \midrule
        SIMP~\cite{bendsoe1988generating} & \cmark & \cmark & \xmark & \xmark & \xmark & \xmark & \cmark & \cmark \\
        TopologyGAN~\cite{nie2021topologygan}       & \cmark & \cmark & \xmark & \xmark & \cmark & \cmark & \cmark & \xmark  \\
        TopoDiff~\citep{maze2022topodiff}          & \cmark & \cmark & \xmark & \xmark & \cmark & \cmark & \cmark & \xmark  \\
        TopoDiff-GUIDED~\citep{maze2022topodiff} & \cmark & \cmark & \xmark & \xmark & \cmark & \cmark & \cmark & \cmark  \\
        TopoDiff-FF (ours)       & \cmark & \cmark & \cmark & \cmark & \xmark & \xmark & \cmark & \xmark  \\
        TopoDiff-FF + SIMP (ours)       & \cmark & \cmark & \cmark & \cmark & \xmark & \xmark & \cmark & \cmark  \\
        \bottomrule
    \end{tabular}
    \caption{Conditioning and guiding variables for different optimization methods and model configurations. All models condition directly or indirectly on loads, boundary conditions, and volume fraction. TopologyGAN, TopoDiff, and TopoDiff-GUIDED have additional conditioning on stress and energy fields, where TopoDiff-FF conditions are on the kernels. TopoDiff-GUIDED, TopoDiff-FF + SIMP, and SIMP itself are guided by a measure of performance.
    }
    \label{tab:conditioning-variables}
\end{table*}
The objective of a generic minimum compliance problem for a mechanical system is to
find the material density distribution $\vx \in \mathbb{R}^n$ that minimizes the
structure’s deformation under the prescribed boundary and
loads condition~\citep{sigmund2013topology, liu2014efficient}. Given a set of design variables $\vx = \{\vx_i\}^n_{i=0}$, where $n$ is the domain dimensionality (in our case $n = 64 * 64$), the minimum compliance problems (Fig.~\ref{fig:intro_to}) can be written as: 
\begin{equation}
\begin{aligned}
\min_{\vx} \quad & c(\vx) = F^T U(\vx) \\
\textrm{s.t.} \quad & v(\vx) = v^T \vx < \bar v\\
  & 0 \leq \vx \leq 1    \\
\end{aligned}
\end{equation}
where the goal is to find the design variables that minimize compliance $c(\vx)$ given the constraints. $F$ is the tensor of applied loads and $U(\vx)$ is the node displacement, solution of the equilibrium equation $K(\vx) U(\vx) = F$ where $K(\vx)$ is the stiffness matrix and is a function of the considered material. $v(\vx)$ is the required volume fraction. The problem is a relaxation of the topology optimization task, where the design variables are continuous between 0 and 1. A simple thresholding mechanism is employed to assign the presence or absence of material in each node in the design domain.
One significant advantage of topology optimization is its ability to create optimized structures that meet specific performance requirements. However, a major drawback of topology optimization is that it can be computationally intensive and may require significant computational resources. Additionally, some approaches to topology optimization may be limited in their ability to generate highly complex geometries and get stuck in local minima. 

\paragraph{Diffusion Models.}
Let $\vx$ denote the observed data which is either continuous $\vx \in \bR^D$ or discrete $\vx \in \{0,...,255\}^D$. Let $\vz_1, ..., \vz_T$ denote $T$ latent variables in $\bR^D$.
We now introduce, the \emph{forward or diffusion process} $q$, the \emph{reverse or generative process} $p_{\theta}$, and the objective $L$.
The forward or diffusion process $q$ is defined as~\cite{ho2020}:
\begin{align}
    q(\vz_{1:T} | \vx) &= q(\vz_1 | \vx) \prod_{t=2}^T q(\vz_t | \vz_{t-1}), \\
    q(\vz_t | \vz_{t-1}) &= \cN(\vz_t | \sqrt{1-\beta_t}~\vz_{t-1}, \beta_t I)
    \label{eq:ddpm_inference_main}
\end{align}
The beta schedule $\beta_1, \beta_2, ..., \beta_T$ is chosen such that the final latent image $\vz_T$ is nearly Gaussian noise.
The generative or inverse process $p_{\theta}$ is defined as:
\begin{align}
    p_{\theta}(\vx, \vz_{1:T}) &= p_{\theta}(\vx | \vz_1) p(\vz_T) \prod_{t=2}^T p_{\theta}(\vz_{t-1}|\vz_t), \\
     p_{\theta}(\vz_{t-1}|\vz_t) &= \cN(\vz_{t-1}|\mu_{\theta}(\vz_t, t), \sigma_t^2 I),
\end{align}
where $p(\vz_T)= \cN(\vz_T| 0, I)$, and $\sigma^2_t$ often is fixed (e.g. to $\sigma^2_t = \beta_t$).
The neural network $\mu_{\theta}(\vz_t, t)$ is shared among all time steps and is conditioned on $t$. 
The model is trained with a re-weighted version of the ELBO that relates to denoising score matching \cite{song2019}.
The negative ELBO $L$ can be written as
\begin{align}
    \bE_q\left[- \log \dfrac{p_{\theta}(\vx, \vz_{1:T})}{q(\vz_{1:T} | \vx)} \right]
    = L_0 + \sum_{t=2}^{T} L_{t-1} + L_T,
    \label{eq:loss_ddpm_main}
\end{align}
where $L_0 = \bE_{q(\vz_1|\vx)} \left[- \log p(\vx |\vz_1) \right]$ is the likelihood term (parameterized by a discretized Gaussian distribution)
and, if $\beta_1,...\beta_T$ are fixed, $L_T = \KL[q(\vz_T|\vx), p(\vz_T)]$ is a constant.
The terms $L_{t-1}$ for $t=2,...,T$ can be written as:

\begin{align}
    L_{t-1} & = \bE_{q(\vz_t|\vx)} \Big[\KL[q(\vz_{t-1}|\vz_t,\vx)\ | \ p(\vz_{t-1}|\vz_t)] \Big] \nonumber \\ 
    & = \bE_{q(\vz_t|\vx)} \left[ \frac{1}{2\sigma_t^2} \|\mu_{\theta}(\vz_t, t) - \tilde{\mu}(\vz_t, \vx)\|_2^2 \right] 
    + C_t \ ,
    \label{eq:Lt_kl}
\end{align}
where $C_t = \frac{D}{2} \left[ \frac{\tilde{\beta}_t}{\sigma_t^2} - 1 + \log \frac{\sigma_t^2}{\tilde{\beta}_t}\right]$.
By further applying the reparameterization trick \cite{kingma2013auto}, the terms $L_{1:T-1}$ can be rewritten as a prediction of the noise $\epsilon$ added to $\vx$ in $q(\vz_t|\vx)$.
Parameterizing $\mu_{\theta}$ using the noise prediction $\epsilon_{\theta}$, we can write
\begin{align}
    \label{eq:Lt_simple}
    L_{t-1, \epsilon} (\vx) &= \bE_{q(\epsilon)} \left[ w_t \|\epsilon_{\theta}(\vz_t(\vx, \epsilon)) -\epsilon\|_2^2 \right],
\end{align}
where $w_t = \frac{\beta_t^2}{2\sigma_t^2\alpha_t (1-\bar{\alpha}_t)}$,
which corresponds to the ELBO objective~\citep{jordan1999introduction}.

\paragraph{Conditioning.} 
Similarly, introducing a conditioning variable $\vc$, conditional modeling can be written as
\begin{equation}
    p_{\vtheta}(\vx, \vz_{1:T} | \vc) = p_{\theta}(\vx | \vz_1, \vc)  p_{\vtheta}(\vz_T | \vc) \prod_{t=2}^T p_{\vtheta}(\vz_{t-1} | \vz_t, \vc),
    \label{eq: conditional-diffusion}
\end{equation}
and we can similarly write the per-layer loss with noise prediction as:
\begin{align}
    \label{eq:Lt_simple}
    L^c_{t-1, \epsilon}(\vx) &= \bE_{q(\epsilon)} \left[ w_t \|\epsilon_{\theta}(
    \vz_t(\vx, \epsilon), \vc) -\epsilon\|_2^2 \right].
\end{align}
We will now discuss different ways to condition the model. 

\paragraph{Guidance.}
Guidance is used to improve sample fidelity vs mode coverage in conditional diffusion models post-training~\cite{dhariwal2021diffusion, ho2021classifierfree}.
Classifier guidance~\cite{dhariwal2021diffusion} is based on bayesian inversion and the fact that inverse problems can be tackled relatively easily in diffusion models. In particular, the goal of the guidance is to shift the model mean $\mu_{\theta}$ closer to the target guidance $\vc$.
The general idea in classifier guidance is to leverage Bayes inversion
\begin{equation}
    p(\vz | \vg) \propto p(\vg | \vz) p(\vz),
\end{equation}
and notice that training the conditional model $p(\vz | \vg)$ is equivalent to training an unconditional model $p(\vz)$ and a classifier $p(\vg | \vz)$. Then the training signal will be a composition of unconditional and conditional scores for the model.
Taking the gradients, we can see that classifier guidance is shifting the unconditional score for each class~\cite{dhariwal2021diffusion}.
We can leverage similar ideas for feasible vs unfeasible designs. With similar reasoning, regression guidance~\cite{maze2022topodiff}, aims to guide the sampling process towards configurations with low regression error as the target. In particular, the regression target for generating topologies is low compliance error between predicted (with a regression model) and real (obtained running a FEM solver) compliance. Guidance is useful to reduce high-compliance configurations generated by the model.

\paragraph{TopoDiff.}
Conditional diffusion models have been adapted for constrained engineering problems with performance requirements.
TopoDiff~\citep{maze2022topodiff} proposes to condition on loads, volume fraction, and fields similarly to~\cite{nie2021topologygan} to learn a constrained generative model (Fig.~\ref{fig:topodiff-intro}). 
In particular, the generative model can be written as: 
\begin{equation}
    p_{\theta}(\vx_{t-1} | \vx_{t}, \vc, \vg) = \mathcal{N}(\vx_{t-1} ~\vert~ \mu_{\theta}(\vx_t, \vc) + \vg_{c} + \vg_{fm}, ~\sigma),
\end{equation}
where $\vc$ is a conditioning term and is a function of the loads $l$, volume fraction $v$, and fields $f$, i.e $\vc = h(l, v, f)$. The fields considered are the Von Mises stress $\sigma_{vm} = (\sigma^2_{11} - \sigma_{11}\sigma_{22} + \sigma^2_{22} + 3\sigma^2_{12})^{1/2}$ and the strain energy density field $W=(\sigma_{11}\epsilon_{11} + \sigma_{22}\epsilon_{22} + 2 \sigma_{12}\epsilon_{12})/2$. 
Here $\sigma_{ij}$ and $\epsilon_{ij}$ are the stress and energy components over the domain. 
$\vg$ is a guidance term, containing information to guide the sampling process toward regions with low floating material (using a classifier and $\vg_{fm}$) and regions with low compliance error, where the generated topologies are close to optimized one (using a regression model and $\vg_c$). Where conditioning $\vc$ is always present and applied during training, the guidance mechanism $\vg$ is optional and applied only at inference time.

\paragraph{Limitations.}
TopoDiff is an effective method for generating topologies that satisfy constraints and have low compliance errors. However, the generative model is computationally expensive as hundreds of layers need to be sampled for each sample. Furthermore, the model requires preprocessing of configurations using a FEM solver, which is also computationally expensive and time-consuming. This approach relies heavily on fine-grained knowledge of the domain input, limiting its applicability to more challenging topology problems.
Additionally, the model requires training of two surrogate models (a classification and regression model) which can be helpful for out-of-distribution configurations. However, to train the regression model, additional optimal and suboptimal topologies are needed assuming access to the desired performance metric on the train set. Similarly, the classifier requires additional labeled data to be gathered.

\section{Method}

\begin{figure}
    \centering
    \includegraphics[width=.9\linewidth]{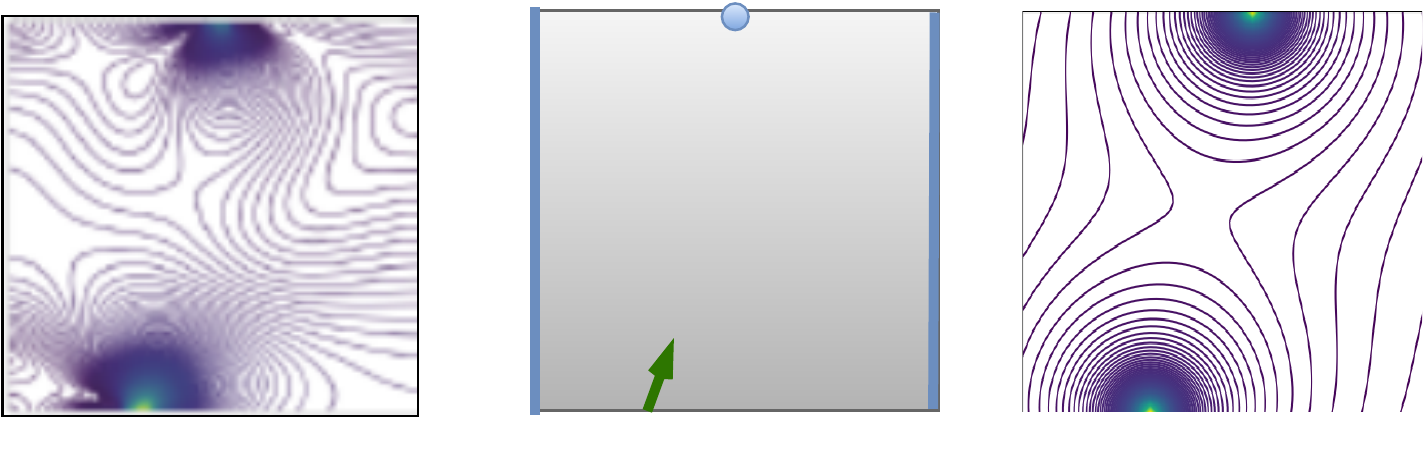}
    \caption{Field (left) vs Kernel (right) conditioning. Where computing the fields requires solving an expensive iterative FEA problem for each new configuration, computing the kernel relaxation is a single-step, computationally inexpensive approximation that does not rely on domain knowledge and scales to any resolution or domain structure.}
    \label{fig:kernel-field}
\end{figure}

\begin{table}
    \centering
    \resizebox{0.99\linewidth}{!}{
    \begin{tabular}{c| c c c}
        Class & Metrics & Goal & Challenging?\\
        \toprule
        Hard-constraint       & Loads Disrespect & Feasibility & No     \\
        Hard-constraint       & Floating Material & Manufacturability & Yes \\
        Soft-constraint       & Volume Fraction   & Min Cost & No    \\
        Functional Performance & Compliance Error & Max Performance & Yes\\
        Modeling Requirements  & Sampling Time & Fast Inference & Yes \\
        \bottomrule
    \end{tabular}
    }
    \caption{Design and Modelling requirements for a constrained generative model for topology optimization. Our goal is to improve the requirements that are challenging to fulfill.
    In this work, we focus on improving Floating Material, reducing Compliance Error, and reducing Sampling Time.}
    \label{tab:table-constraints}
\end{table}

Our three main objectives are: improving inference efficiency, improving the sampling time for diffusion-based topology generation while still satisfying the design requirements with a minimum decrease in performance; minimizing reliance on force and strain fields as conditioning information, reducing the computation burden at inference time and the need for ad-hoc conditioning mechanisms for each problem and domain; merging together learning-based and optimization-based methods, refining the topology generated using a conditional diffusion model, improving the final solution in terms of manufacturability and performance. 
We aim to improve generative models with constraints and eliminate the need for costly FEM solutions as conditioning mechanisms.
We refer to our approach as TopoDiff-FF, short for Topology-Diffusion-Fields-Free, and to our approach with optimization-based refinement as TopoDiff-FF + SIMP.

\paragraph{Conditioning on Hard and Soft Constraints.}
All models are subject to conditioning based on loads, boundary conditions, and volume fractions. In addition, TopoDiff and TopoDiff-GUIDED undergo conditioning based on force field and energy strain, while TopoDiff-FF is conditioned based on kernel relaxation, which is discussed in the following paragraph (see Fig.~\ref{fig:kernel-field}).

\paragraph{Green's Functions.}
To improve the efficiency of diffusion-based topology generation and minimize reliance on force and strain fields, we aim to relax boundary conditions and loads by leveraging kernels as approximations for the way such constraints act on the domain. One possible choice of kernel structure is inspired by Green's method~\cite{garabedian1960partial}, which defines integral functions that are solutions to the time-invariant Poisson's Equation~\cite{hale2013introduction}, a generalization of Laplace's Equation for point sources excitations.
Poisson's Equation can be written as $\nabla^2_x f(x) = h$, where $h$ is a forcing term and $f$ is a generic function defined over the domain $\mathcal{X}$. This equation governs many phenomena in nature, and a special case is a forcing part $h=0$, which yields the Laplace's Equation formulation commonly employed in heat transfer problems.
Green's method is a mathematical construction to solve partial differential equations without prior knowledge of the domain. The solutions obtained with this method are known as Green's functions~\cite{keldysh1951characteristic}. While solutions obtained with this method can be complex in general, for a large class of physical problems involving constraints and forces that can be approximated with points, a simple functional form can be derived by leveraging the idea of source and sink.
Consider a laminar domain (e.g., a beam or a plate) constrained in a feasible way. If a point source is applied to this domain (e.g., a downward force on the edge of a beam or on the center of a plate) in $x_f$, such force can be described using the Dirac delta function, $\delta(x - x_f)$. The delta function is highly discontinuous but has powerful integration properties. In particular $\int f(x) \delta(x - x_f) dx = f(x_f)$ over the domain $\mathcal{X}$. The solution of the time-invariant Poisson's Equation with point concentrated forces can be written as a Green's function solution, where the solution depends only on the distance from the force application point. In particular:

\begin{equation}
        \mathcal{G}(x, x') = - \dfrac{1}{4 \pi} \dfrac{1}{|x - x'|},
    \end{equation}
where $r = |x - x'| = \sqrt{|x_i - x^{'}_{i}|^2 + |x_j - x^{'}_{j}|^2}$.
We propose to approximate the forces and loads applied to our topologies using a kernel relaxation built using Green's functions. While this formulation may not provide a correct solution for generic loads and boundary conditions, it allows us to provide computationally inexpensive conditioning information that respects the original physical and engineering constraints. By leveraging these ideas, we aim to increase the amount of information provided to condition the model, ultimately improving generative models with constraints.

\paragraph{Kernel Relaxation.}
We can use these kernels to construct a kernel relaxation method to condition generative models (Fig.~\ref{fig:kernel-field}). The idea is to use the kernels as approximations of the way boundary conditions and loads act on the domain. Specifically, we can use the kernels to represent the effects of the boundary conditions and loads as smooth functions across the domain. This approach avoids the need for computationally expensive and time-consuming finite element Analysis (FEA) to provide conditioning information.
To apply the kernel relaxation method, we first determine the locations of the sources and sinks in the domain. Then, we compute the corresponding kernels for the loads and boundary conditions, respectively. Finally, we use the resulting kernels as conditioning information for generative models.
In particular, we consider loads as sources and boundary conditions as sinks. 
For a load or source $p$, and $r = |x - x^p| = \sqrt{|x_i - x^{p}_{i}|^2 + |x_j - x^{p}_{j}|^2}$ we have:
\begin{equation}
    K_{l}(x, x^p; \alpha, \beta) \propto (1 - e^{- \alpha/r^{\beta}}) ~ \bar p,
\end{equation}
where $\bar p$ is the module of a generic force in 2D.
Notice how, for $r \rightarrow 0$, $K_l(x, x^p) \rightarrow p$, and $r \rightarrow \infty$, $K_l(x, x^p) \rightarrow 0$.
For a boundary condition or sink:
\begin{equation}
    K_{bc}(x, x^{bc}; \alpha, \beta) \propto e^{- \alpha/r^{\beta}}.
\end{equation}
We notice how closer to the boundary the kernel is null, and farther from the boundary the kernel tends to 1.
Note that the choice of $\alpha$ and $\beta$ parameters in the kernels affect the smoothness and range of the kernel functions. By adjusting these parameters, we can control the trade-off between accuracy and computational cost. Furthermore, these kernels are isotropic, meaning that they do not depend on the direction in which they are applied.
Overall, the kernel relaxation method offers a computationally inexpensive way to condition generative models on boundary conditions and loads, making them more applicable in practical engineering and design contexts.

\section{Experiments}

\begin{figure}
    \centering
    \includegraphics[width=.95\columnwidth]{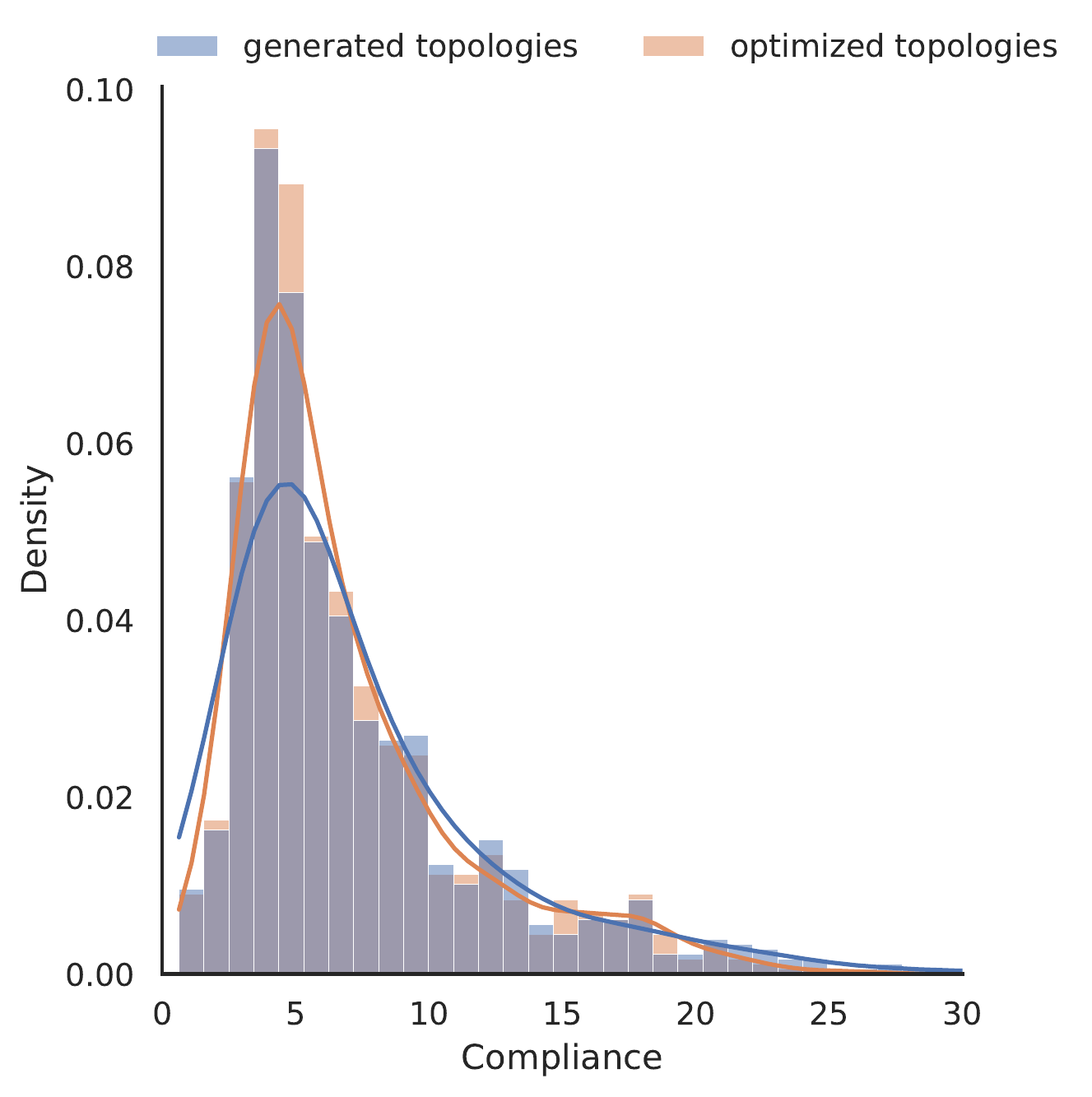}
    \caption{Histogram empirical distribution generated and optimized compliances for task-2 (unknown constraints). We see that the generated topologies match well the distribution of compliance for unseen constraint configurations.
    }
    \label{fig:distro-compliance}
\end{figure}

\paragraph{Setup.}
We train all the models for 200k steps on 30000 optimized topologies on a 64x64 domain. 
We set the hyperparameters, conditioning structure, and training routine as proposed in~\citep{maze2022topodiff}.
For all the models we condition on volume fraction and loads.
For TopoDiff we condition on additional stress and energy fields.
For TopoDiff-FF we condition on boundary conditions and kernels.
TopoDiff-GUIDED leverages a compliance regressor and floating material classifier guidance (CLS+REG).
We use a reduced number of sampling steps for all the experiments. See Table~\ref{tab:table-constraints} for an overview of conditioning and guidance variables used for different models and methods considered in this work.

\paragraph{Dataset.}
We use the dataset of optimized topologies gathered using SIMP proposed in~\cite{maze2022topodiff}.
\begin{table}[ht]
    \centering
    \begin{tabular}{c|cccc}
         Cut-off & 10     & 25     & 50     & 100    \\
         \toprule
         $F(x)$  & 0.9871 & 0.9963 & 0.9985 & 0.9990 \\
    \end{tabular}
    \caption{Cumulative Density Function for compliance cut-off on the train set. We can see that more than 99.90\% of the training set has compliance lower than 100.}
    \label{tab:t-F(t)}
\end{table}
For each topology, we have information about the loading condition, boundary condition, volume fraction, and optimal compliance. Additionally, for each configuration, a pre-processing step computes the force and strain energy fields (see Fig.~\ref{fig:kernel-field}). 

\paragraph{Evaluation.}
We employ both engineering and generative metrics to evaluate the effectiveness of our model. Our metrics encompass physical, engineering, and modeling requirements to determine the model's overall performance.
We measure the Volume Fraction Error (VFE), which calculates the deviation between the generated topology volume and the prescribed volume. We also use Floating Material (FM) to assess the manufacturability of the model.
For performance, we use the structural compliance metric~\cite{sigmund2013topology}, which measures the overall displacement under specified constraints. Minimizing compliance is one of the primary objectives of topology optimization. We calculate the average and median compliance (Avg C, Mdn C) for all tasks.
We also use the Compliance Error (CE) to evaluate the engineering performance, which determines the deviation between the compliance calculated by the SIMP optimization method and the compliance generated by the diffusion model. Additionally, we analyze the inference time, which includes preprocessing and sampling time, to determine the model's speed in creating fast concept designs.
To assess the performance of the generative models, we design three test tasks: task-1, task-2, and task-3. We divide the test configurations into various groups based on their constraints and expected optimal performance. Table~\ref{tab:task} presents these tasks.
We notice that all the models have null load disrespect (LD), showing that fulfilling this hard constraint is easy for deep generative models.

\begin{itemize}
    \item Task-1: the constraints in the task-1 test set are identical to those in the training set. We filter out generated configurations with high compliance when measuring performance on this task.
    \item Task-2: the constraints in the task-2 test set differ from those in the training set. We filter out generated configurations with high compliance when measuring performance on this task.
    \item Task-3: the constraints in the task-3 test set are the same as those in task-2. We consider all the generated configurations with low and high compliance when measuring performance on this task.
\end{itemize}

The purpose of these tasks is to evaluate the generalization capability of the machine learning models in- and out-of-distribution. By testing the models on different test sets with varying levels of difficulty, we can assess how well the models can perform on new, unseen data.
More importantly, we want to understand how important the role of the force field and energy strain is with unknown constraints.

\begin{table}[ht]
    \centering
    \begin{tabular}{c | c c c c}
        task & constraints & performance & avg C & mdn C \\
        \midrule
        task-1 & in-distro & in-distro  & 7.73 & 5.54  \\
        task-2 & out-distro & in-distro & 6.84 & 5.29  \\
        task-3 & out-distro & out-distro & 6.85 & 5.31 \\
    \end{tabular}
    \caption{Tasks in order of challenge for the model. 
    We expect the models to perform well on task-1, acceptably on task-2, and fail on task-3. We also report the average and median compliance over the different sets as a measure of optimality for the generated topologies.}
    \label{tab:task}
\end{table}

\subsection{Performance with In-Distribution Constraints}
\label{subsection:in-distro}
\begin{table*}[ht]
    \centering
    \setlength{\tabcolsep}{3pt}
    \resizebox{0.99\textwidth}{!}{
    \begin{tabular}{c|  c c c c c  c c c c c c c}
        & Task & Steps  & Constraints & Guidance & Avg C $\downarrow$ & Mdn C $\downarrow$ & \%~CE $\downarrow$ & \%~VFE $\downarrow$ & \%~FM $\downarrow$ & Sampling $\downarrow$ & Processing $\downarrow$ &  Inference $\downarrow$ \\
        \midrule
        TopoDiff              & task-1  & 100 & FIELD  & COND   & 8.01 & 5.60  & \underline{5.46} & \underline{1.47}  & \underline{5.79} & 2.23 & 3.31 & 5.54        \\
        TopoDiff-GUIDED       & task-1  & 100 & FIELD  & CLS+REG & 8.15 & 5.63 & 5.93 & 1.49 & 5.82  &  2.46 & 3.31 & 5.87 \\
        TopoDiff-FF           & task-1  & 100 & KERNEL & COND    & 9.76 & 6.31 & 24.90 & 2.05 & 8.15 &  2.23 & 0.12 & \textbf{2.35} \\
        TopoDiff-FF + SIMP (10) & task-1  & 100 & KERNEL & COND & 8.05 & 5.77 & \textbf{4.16} & \textbf{1.16} & \textbf{5.61} & 2.23 & 0.32 & \underline{2.55} \\
        \bottomrule
    \end{tabular}
    }
    \caption{Evaluation of different model variants on in-distribution constraints. We remove samples with compliance >100.
    C: compliance.
    CE: compliance error. VFE: volume fraction error. FM: floating material. We use 100 sampling steps for all models. SIMP (10) means that we run SIMP for 10 iterations. Results averaged over 5 runs.
    For a general overview with confidence intervals see the left side of Fig.~\ref{fig:barplots}.
    }
    \label{tab:constraints-in-distro}
\end{table*}
Table~\ref{tab:constraints-in-distro} presents an evaluation of different model variants on in-distribution constraints for Task 1, with various measures of compliance, compliance error, volume fraction error, and floating material. Lower is better for all the metrics.
The models evaluated include TopoDiff, TopoDiff-GUIDED, TopoDiff-FF, and TopoDiff-FF + SIMP (10), with varying types of constraints and guidance. 
TopoDiff and TopoDiff-GUIDED condition of stress and energy fields, with TopoDiff-GUIDED leveraging surrogate models for guidance (classifier and regressor); TopoDiff-FF conditions on kernel relaxation.
The table shows that TopoDiff and TopoDiff-GUIDED have lower values for most measures of compliance, compliance error, volume fraction error, and floating material compared to TopoDiff-FF. 
The TopoDiff-FF model has higher compliance error as expected but comparable VFE and FM, indicating that a lightweight kernel relaxation is a viable conditioning option.
Additionally, TopoDiff-FF + SIMP (10) has the lowest values for compliance error, volume fraction error, and floating material.
The table also reports the time taken for sampling, pre-processing, and inference for each model. TopoDiff-FF, even reducing the number of sampling steps for TopoDiff to 100, has significantly lower times for pre-processing and overall inference, because of the absence of FEA pre-processing.
Overall, the results suggest that the TopoDiff and TopoDiff-GUIDED models perform better than TopoDiff-FF on in-distribution constraints, with TopoDiff-FF (and TopoDiff-FF+SIMP) being much faster at sampling and a viable data-driven alternative when FEA is impractical or computationally unfeasible.
The results for TopoDiff-FF+SIMP suggest that using a combination of deep generative models and a simplified iterative method (SIMP) can lead to better results in terms of compliance and design requirements, even if the gap with TopoDiff is not so relevant when dealing with the in-distribution scenario.

\subsection{Performance with Out-of-Distribution Constraints}
\begin{figure*}[ht]
    \centering
    \includegraphics[width=.95\linewidth]{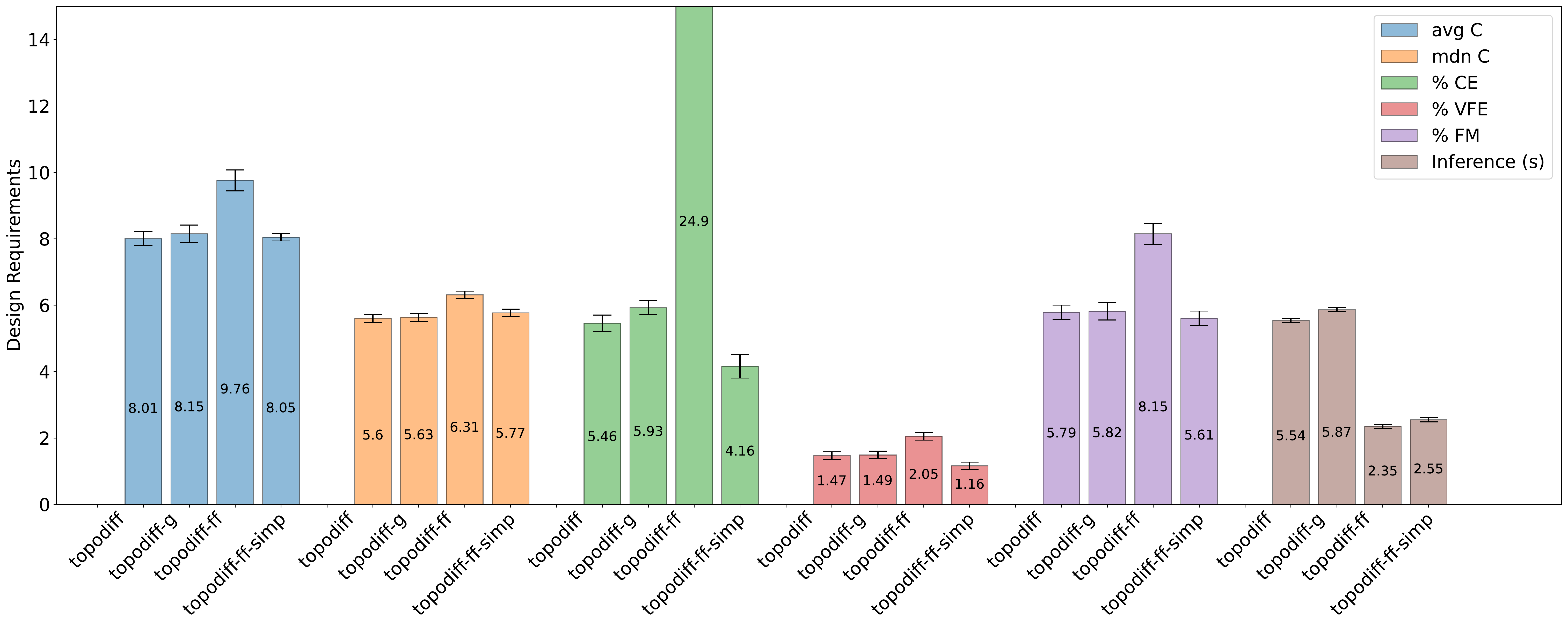}
    \includegraphics[width=.95\linewidth]{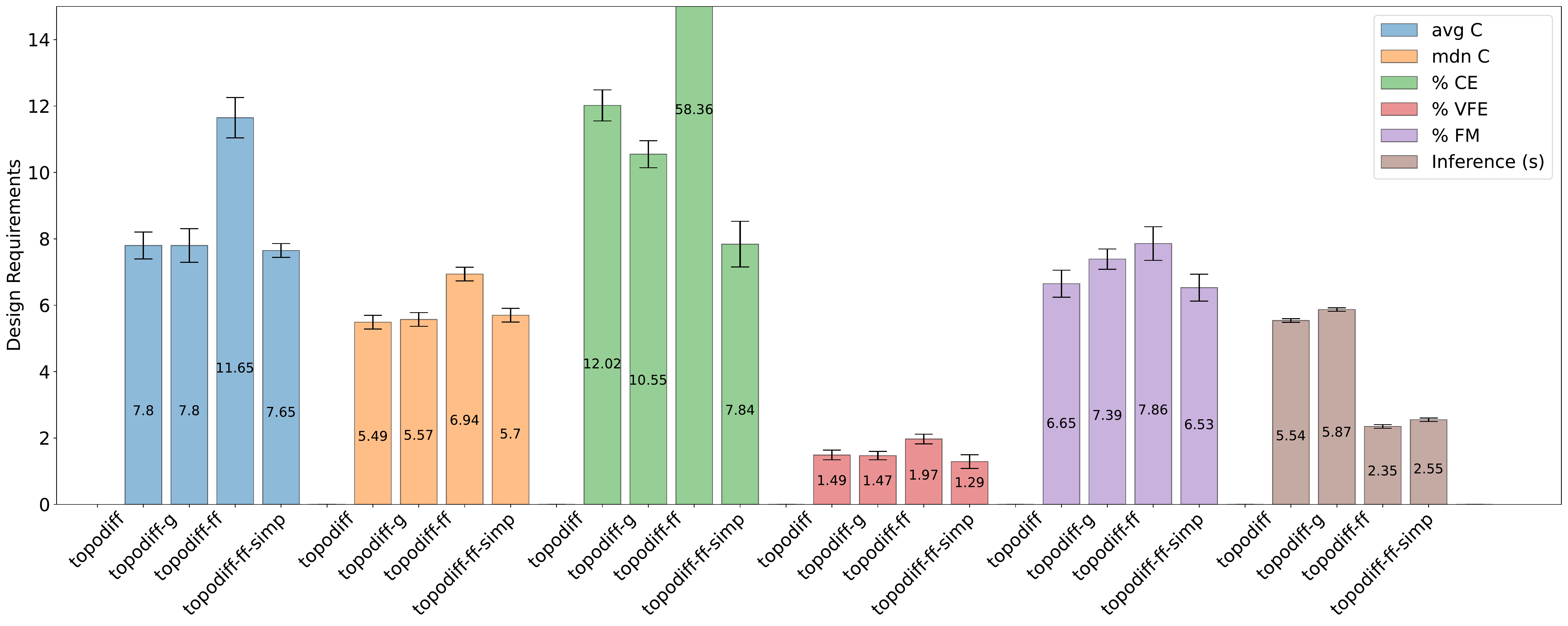}
    \caption{Overview design requirements for the considered models for task-1 (in-distribution constraints, top) and task-2 (out-of-distribution constraints, bottom).
    From left to right: average compliance in blue. Median compliance in orange. Compliance error in green. Volume fraction error in red. Floating material in purple. Inference time in brown. 
    Lower is better for all the metrics.}
    \label{fig:barplots}
\end{figure*}

\begin{table*}[ht]
    \centering
    \setlength{\tabcolsep}{3pt}
    \resizebox{0.99\textwidth}{!}{
    \begin{tabular}{c|c c c c c c c c c c c c}
        & Task & Steps & Constraints & Guidance & Avg C $\downarrow$ & Mdn C $\downarrow$  & \%~CE $\downarrow$  & \%~VFE $\downarrow$ & \%~FM $\downarrow$ & Sampling $\downarrow$ & Processing $\downarrow$ &  Inference $\downarrow$ \\
        \midrule
        TopoDiff            & task-2 & 100 & FIELD & COND & 7.80 & 5.49 & 12.02 &  1.49 & \underline{6.65} & 2.23 & 3.31  & 5.54      \\
        TopoDiff-GUIDED     & task-2 & 100 & FIELD    & CLS+REG & 7.80 & 5.57 & \underline{10.55}  & \underline{1.47} & 7.39 & 2.46  & 3.31 & 5.87 \\
        TopoDiff-FF         & task-2 & 100 & KERNEL & COND & 11.65 & 6.94 & 58.36 &  1.97 & 7.86 & 2.23  & 0.12 & \textbf{2.35}  \\
        TopoDiff-FF + SIMP (10)    & task-2 & 100 & KERNEL & COND & 7.65 & 5.70 & \textbf{7.84} & \textbf{1.29} & \textbf{6.53} & 2.23  & 0.32 & \underline{2.55} \\
        \bottomrule
    \end{tabular}
    }
    \caption{Evaluation of different model variants on out-of-distribution constraints. We remove samples with compliance >100.
    C: compliance.
    CE: compliance error. VFE: volume fraction error. FM: floating material. We use 100 sampling steps for all models. SIMP (10) means that we run SIMP for 10 iterations. Results averaged over 5 runs.
    For a general overview with confidence intervals see the right side of Fig.~\ref{fig:barplots}.}
    \label{tab:constraints-out-distro}
\end{table*}
Table~\ref{tab:constraints-out-distro} reports results on task-2 with out-of-distribution constraints. The considered models, conditioning, and guidance mechanisms are the same as explained in Subsec.~\ref{subsection:in-distro}.
The results show that in out-of-distribution scenarios, TopoDiff-GUIDED model performs better than the TopoDiff, achieving a lower CE and VFE. TopoDiff-FF performs well in terms of VFE and FM, but poorly in terms of compliance, with high CE. However, the TopoDiff-FF + SIMP (10) model significantly improves the performance on Task-2, achieving the lowest CE, VFE, and FM values among all models.
As for task-1, the inference time for TopoDiff-FF and TopoDiff-FF+ SIMP is much better than for the full model.
Overall, the table shows that the use of guidance and optimization techniques, such as SIMP, can significantly improve the performance of topology optimization models, especially on out-of-distribution constraints. See Fig~\ref{fig:barplots} for a visual overview with confidence intervals over five runs.

\subsection{Performance with Out-of-Distribution Constraints and Outliers}
Table~\ref{tab:constraints-out-distro-out-performance} presents results on task-3 with out-of-distribution and without filtering out outliers, i.e configurations with high-compliance (low performance).
High-compliance configurations (>50/100) are out-of-distribution performance for the training set compliance distribution. In particular in Table~\ref{tab:t-F(t)} where we show that more than 99.8\% of the configurations have compliance lower than 50. This means that when we obtain generated samples with such high compliance, the model has partially failed to fulfill the design constraints.
The considered models, conditioning, and guidance mechanisms are the same as explained in Subsec.~\ref{subsection:in-distro}.
\begin{table*}[ht]
    \centering
    \setlength{\tabcolsep}{3pt}
    \resizebox{0.99\textwidth}{!}{
    \begin{tabular}{c|c c c c c c c c c c c c c}
        & Task & Steps & Constraints & Guidance & Avg C $\downarrow$ & Mdn C $\downarrow$  & \%~CE $\downarrow$ & \%~VFE $\downarrow$ & \%~FM $\downarrow$ & Sampling $\downarrow$ & Processing $\downarrow$ &  Inference $\downarrow$ \\
        \midrule
        TopoDiff              & task-3 & 100 & FIELD   & COND    & 8.91 & 5.52 & 31.04 & 1.49 & \underline{6.61} & 2.23 & 3.31 & 5.54  \\
        TopoDiff-GUIDED       & task-3 & 100 & FIELD    & CLS+REG & 8.23 & 5.58 & \underline{17.69}  & \underline{1.47} & 7.36 & 2.46 & 3.31 & 5.87 \\
        TopoDiff-FF & task-3 & 100 & KERNEL & COND    & 19.01 & 7.22 & 144.23 &  2.00 & 8.70  & 2.23 & 0.12 & \textbf{2.35}  \\
        TopoDiff-FF + SIMP (10)  & task-3 & 100 & KERNEL & COND & 7.65 & 5.70 & \textbf{7.84} & \textbf{1.29} & \textbf{6.53} & 2.23  & 0.32 & \underline{2.55}  \\
        \bottomrule
    \end{tabular}
    }
    \caption{Evaluation of different model variants on out-of-distribution constraints. We do not filter out samples with high compliance.
    C: compliance.
    CE: compliance error. VFE: volume fraction error. FM: floating material. We use 100 sampling steps for all models. SIMP (10) means that we run SIMP for 10 iterations. Results averaged over 5 runs.}
    \label{tab:constraints-out-distro-out-performance}
\end{table*}
The results in this table are similar to task-2, where TopoDiff-GUIDED performs better than TopoDiff in terms of CE and TopoDiff-FF achieves high compliance. However, using TopoDiff-FF+SIMP we not only improve all the metrics, but we completely get rid of such outlier configurations, being easy for the optimization process to fix the generated high-compliance configurations in a few iterations. Contrarily, for all other methods, there are always outliers in terms of topologies with high compliance that cannot easily be fixed, making a strong case for unifying deep generative models and optimization for topology optimization in challenging scenarios.

\subsection{Compliance Analysis}
\begin{figure}[ht]
    \centering
\includegraphics[width=.85\columnwidth]{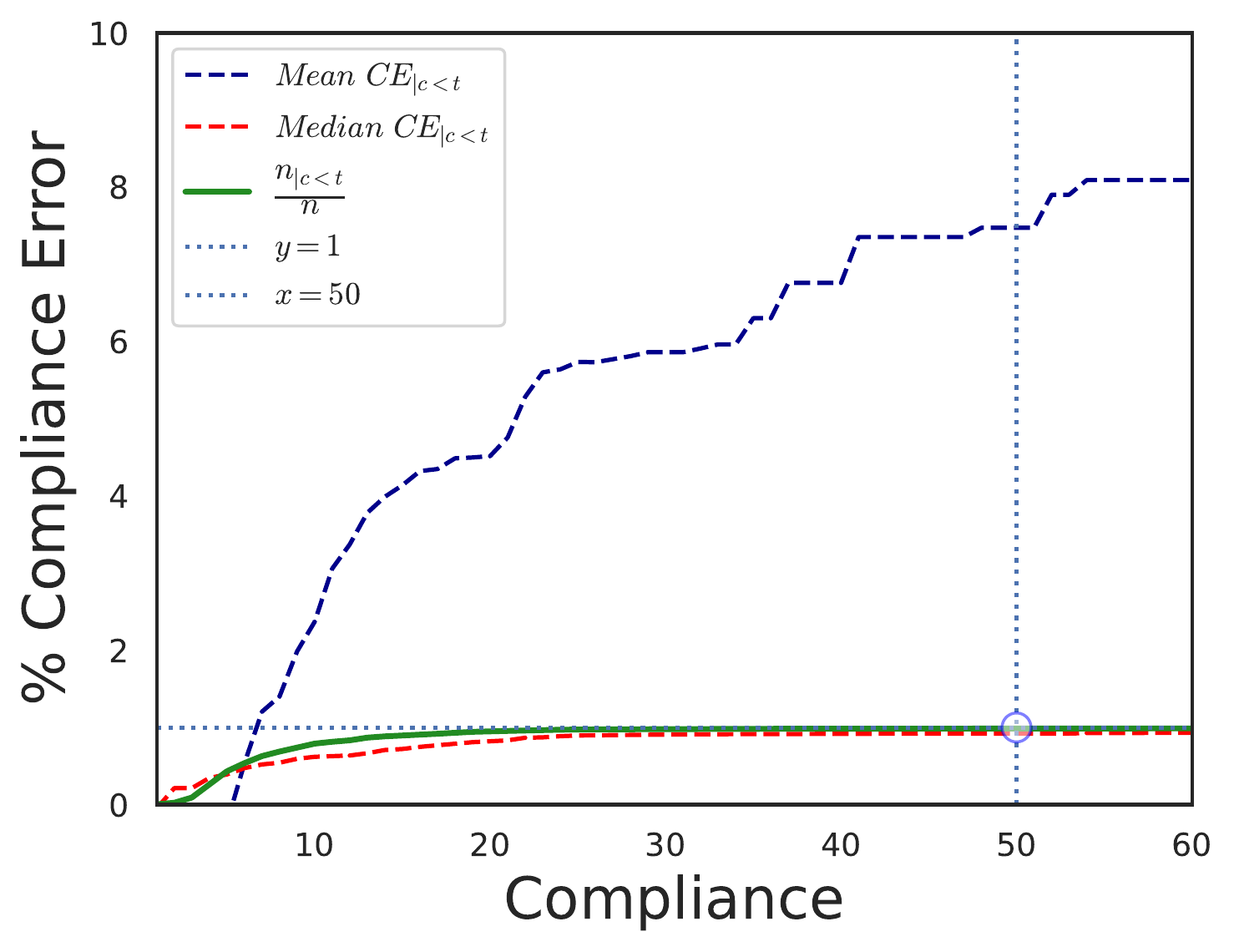}
    \caption{Mean and Median Compliance error for different compliance thresholds.
    We see that more than 99.8~\% of the data has compliance lower than 50 and how to mean compliance error increases with compliance, where the median compliance error plateau. }
    \label{fig:compliance-threshold}
\end{figure}

\begin{figure}[ht]
    \centering
\includegraphics[width=.85\columnwidth]{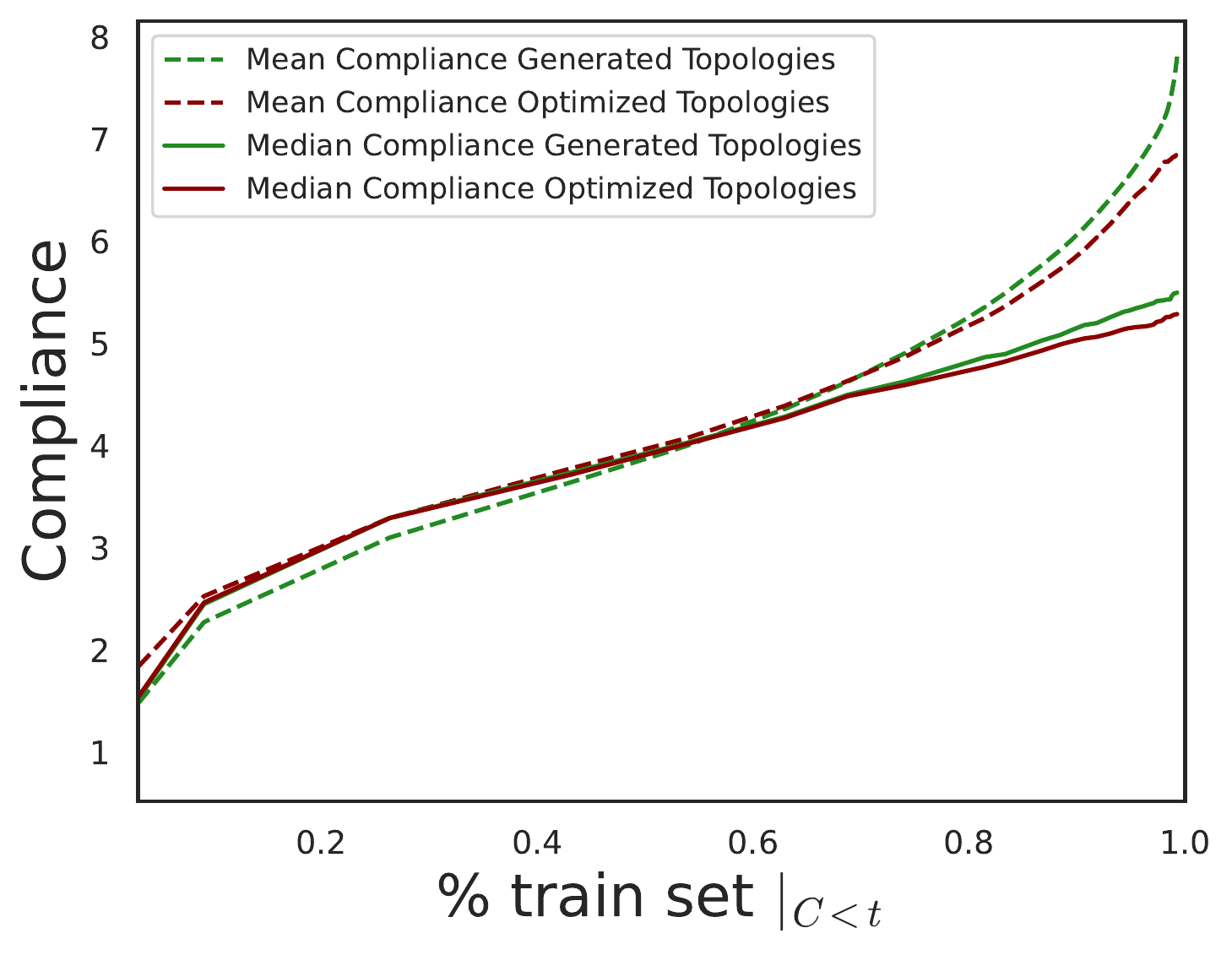}
\quad%
    \caption{Mean and Median Compliance vs \% train set for task-2 (unknown) constraints.
    The generated topologies using TopoDiff perform well for low compliance values.
    Most of the loss in terms of compliance happens for high compliance values.
    }
    \label{fig:compliance-mean-median-level2}
\end{figure}

Here we discuss and study how compliance influences the generated topology quality.
Compliance magnitude seems to be the most important factor for the generation quality and manufacturability of proposed designs.
In Fig.~\ref{fig:distro-compliance} we show qualitative information, providing empirical compliance distribution for generated vs optimized topologies. In Fig.~\ref{fig:compliance-mean-median-level2} and Fig.~\ref{fig:compliance-threshold}
The growth of compliance error with compliance indicates that the model struggles to generate high-compliance topologies that fulfill the design requirements. This is further supported by the fact that very few configurations have high compliance, as indicated by the dataset. These findings highlight the difficulty of modeling compliant structures with high accuracy and precision and suggest that additional techniques or modifications to the model may be necessary to address this challenge. Moreover, the limited ability of the model to generate high-compliance topologies could have important implications for real-world applications, where compliant structures are often critical components of engineering designs. Therefore, improving the model's ability to generate high-compliance topologies is an important research direction for advancing the state-of-the-art in this field.

\begin{figure*}[ht]
    \centering
    \includegraphics[width=.9\linewidth]{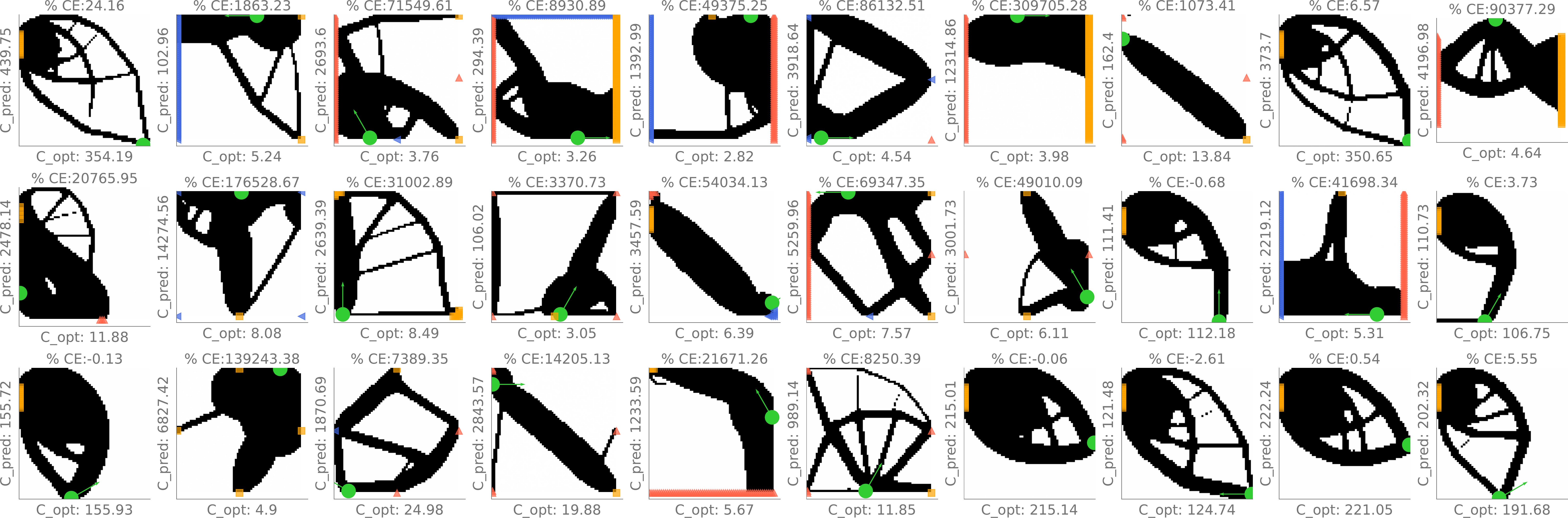}
    \caption{Examples of generated topologies with bad performance.}
    \label{fig:high-compliance}
\end{figure*}

\paragraph{High-Compliance Configurations.}
High-compliance configurations are often associated with mechanically unstable structures, meaning that they can collapse or deform easily under load. As a result, practical engineering designs often prioritize achieving low-compliance solutions that are stable, efficient, and lightweight. However, it is still important to study high-compliance configurations as they can provide valuable insights into the behavior and limitations of structural systems, as well as inspire new design concepts and approaches. We show some examples of high-compliance and high-compliance errors in Fig.~\ref{fig:high-compliance}. For low-compliance and low-compliance errors, see Fig.~\ref{fig:low-compliance}.

\subsection{Inference Time}
Where in~\cite{maze2022topodiff} has been shown that TopoDiff-based models largely outperform topologyGAN-based models~\cite{nie2021topologygan} on design performances (CE, VFE, FM) for in- and out-distribution scenarios, inference time for GAN-based model is still much better than for Diffusion-based models. 
\begin{figure}[ht]
\includegraphics[width=.85\columnwidth]{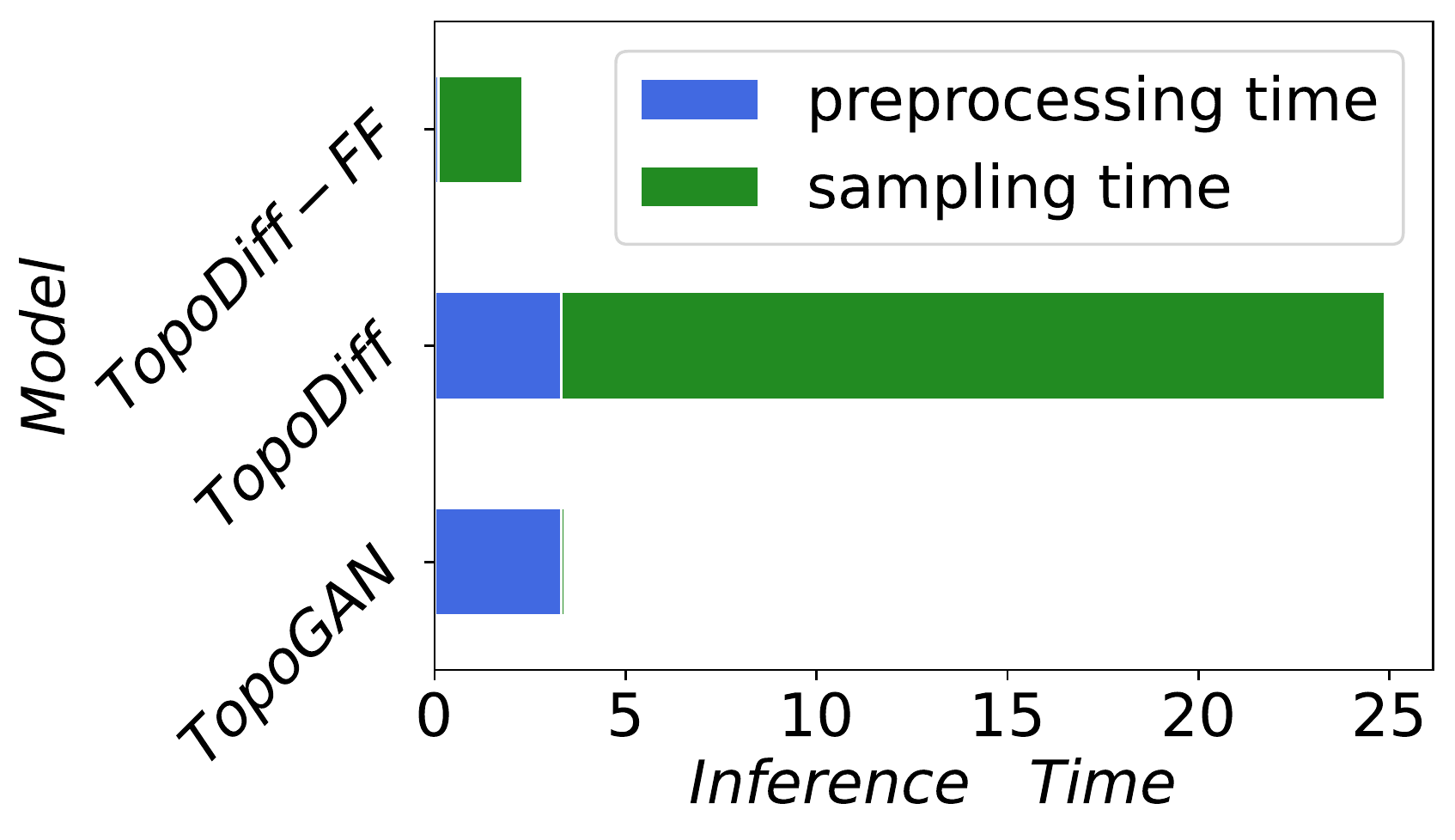}
    \caption{Sampling time. In this figure, we compare the total inference time (pre-processing and sampling time) for TopoDiff, TopologyGAN, and TopoDiff-FF. We can see that we completely remove the pre-processing time component using the kernel relaxation. And reducing the number of sampling steps reduces, even more, the sampling time, making TopoDiff-FF relatively fast at the price of generating configurations with higher compliance than TopoDiff and TopoDiff-GUIDED.}
    \label{fig:sampling}
\end{figure}
Here we present our approach to improving the efficiency of inference for TopoDiff-FF, measured in terms of both pre-processing and sampling time. Specifically, we focus on two key areas of optimization:
I) Reducing the number of steps required for sampling to improve overall sampling time.
II) Leveraging the kernel relaxation technique to reduce pre-processing time by avoiding the computation of force and energy fields.
To demonstrate the effectiveness of our approach, we compare the pre-processing and sampling times of TopoGAN, TopoDiff, and TopoDiff-FF in Fig.~\ref{fig:sampling}. Our results show that TopoDiff-FF is able to generate high-quality samples that satisfy prescribed constraints and exhibit strong performance, all while requiring significantly less time for both pre-processing and sampling compared to TopoDiff and slightly less time than TopoGAN. This improved efficiency is a key goal of TopoDiff-FF, as it simplifies the conditioning process and enables faster inference.

\subsection{Measuring the Relative Gap in Performance}
To better summarize the results on in-distribution and out-of-distribution constraints, we propose to evaluate the models in terms of the relative design gap between the models, using a global metric that accounts for all the requirements presented in Table~\ref{tab:table-constraints}.
We also consider relative ranking (Avg Rank) between the considered models.
The gap and the rank are global metrics that account for compliance, compliance error, volume fraction error, floating material, processing, and inference time. For the average gap (Avg Gap), we use average compliance and compliance error. For the median gap (Mdn Gap), the corresponding median values.
The design gap is a proxy to quantify the distance from an optimal design, where all the requirements are perfectly satisfied, we have compliance error null, and inference time is negligible.
Table~\ref{tab:table-constraints} presents a comparison of different conditioning configurations for generative designs in terms of their performance on task-1 and task-2. The table also shows whether the configuration includes kernels and field constraints.
Overall, the models perform similarly in terms of the average and median gap on task-1, with TopoDiff-FF+SIMP (10) performing the best overall. On task-2, TopoDiff-FF loses in terms of the average gap because of the large CE on out-of-distribution. Interestingly, when considering average ranking, the gap is reduced because of the fast processing and inference time for TopoDiff-FF.
As noted with the previous experiments, TopoDiff-FF with SIMP outperforms all other models in terms of both design quality and relative performance on both tasks, which suggests that the use of SIMP in conjunction with deep generative models is a promising approach for generating high-quality designs for topology optimization.

\begin{table}[ht]
    \centering
    \setlength{\tabcolsep}{3pt}
    \resizebox{0.99\columnwidth}{!}{
    \begin{tabular}{c| c c c c c c c c c c c}
                    & Task & Kernels  & Fields & Mdn Gap $\downarrow$ & Avg Gap $\downarrow$ & avg Rank $\downarrow$ \\
        \midrule
        TopoDiff           & task-1   & \xmark & \cmark  & 3.12  & 4.38 & 2.2\\
        TopoDiff-GUIDED    & task-1   & \xmark & \cmark & 3.16  & 4.52 & 3.2\\
        TopoDiff-FF        & task-1   & \cmark & \xmark  & 4.37  & 7.86 & 2.8\\
        TopoDiff-FF + SIMP & task-1   & \cmark & \xmark  & \textbf{2.69}  & \textbf{3.57} & \textbf{1.6} \\
\midrule
        TopoDiff             & task-2      & \xmark & \cmark  &  3.35 & 5.58  & 2.2  \\
        TopoDiff-GUIDED      & task-2      & \xmark & \cmark  &  3.55 & 5.54  & 3.2 \\
        TopoDiff-FF          & task-2      & \cmark & \xmark  &  5.86 & 13.69 & 2.8 \\
        TopoDiff-FF + SIMP   & task-2      & \cmark & \xmark  &  \textbf{3.08} & \textbf{4.27}  & \textbf{1.6} \\
    \bottomrule
    \end{tabular}
    }
    \caption{Global overview of different conditioning configurations showing how far the generative designs are from an ideal optimal design. On in-distribution configurations, our approach is extremely effective. On out-of-distribution configuration there is still a gap in performance. Adding few steps of SIMP (10 steps) as refinement is effective in improving the design quality.}
    \label{tab:design-gap}
\end{table}

\subsection{Merging Deep Generative Models and Optimization}
Table~\ref{tab:model-simp} shows the results of experiments conducted with three algorithms, SIMP, TopoDiff-FF, and TopoDiff-FF + SIMP, and tested on a particular data split. The table presents the values of two metrics, average C and Compliance Error, for each algorithm, with varying numbers of iterations. It is clear that increasing the number of iterations resulted in improved performance for all algorithms. Moreover, the combination of TopoDiff-FF and SIMP performed better than the other algorithms, achieving the lowest values of average C and CE in both experiments. These results suggest that using the combination of TopoDiff-FF and SIMP is a way to impose the performance constraints in the model without the need for surrogate models or guidance. Also, we are sure that some physical properties are respected, reducing the FM, VFE, and CE using a fast and relatively cheap refinement (5/10 iterations).

\begin{table}[ht]
    \centering
    \small
    \begin{tabular}{c|c c c c }
    \setlength{\tabcolsep}{3pt}
        & Task & Iter & Avg C $\downarrow$ &  \% CE $\downarrow$ \\
        \midrule
        SIMP                    & task-2  & +5 & 9.87           & 35.13 \\
        TopoDiff-FF     & task-2  & -  & 19.01          & 58.36 \\
        TopoDiff-FF + SIMP      & task-2  & +5 & \textbf{8.67}  & \textbf{20.34} \\
        \midrule
        SIMP                    & task-2  & +10 & 8.57           &  17.61 \\
        TopoDiff-FF     & task-2  & -   & 19.01          & 58.36  \\
        TopoDiff-FF + SIMP      & task-2  & +10 & \textbf{7.65}  & \textbf{7.84} \\
        \midrule
    \end{tabular}
    \caption{Comparison of SIMP, TopoDiff-FF and TopoDiff-FF + SIMP. We see that by merging together the topology generated using a conditional diffusion model and SIMP for refinement we obtain the best results in terms of average compliance and compliance error with just 5 or 10 SIMP iterations.}
    \label{tab:model-simp}
\end{table}

\subsection{Manufacturability}
\begin{figure}[ht]
    \centering
\includegraphics[width=.85\columnwidth]{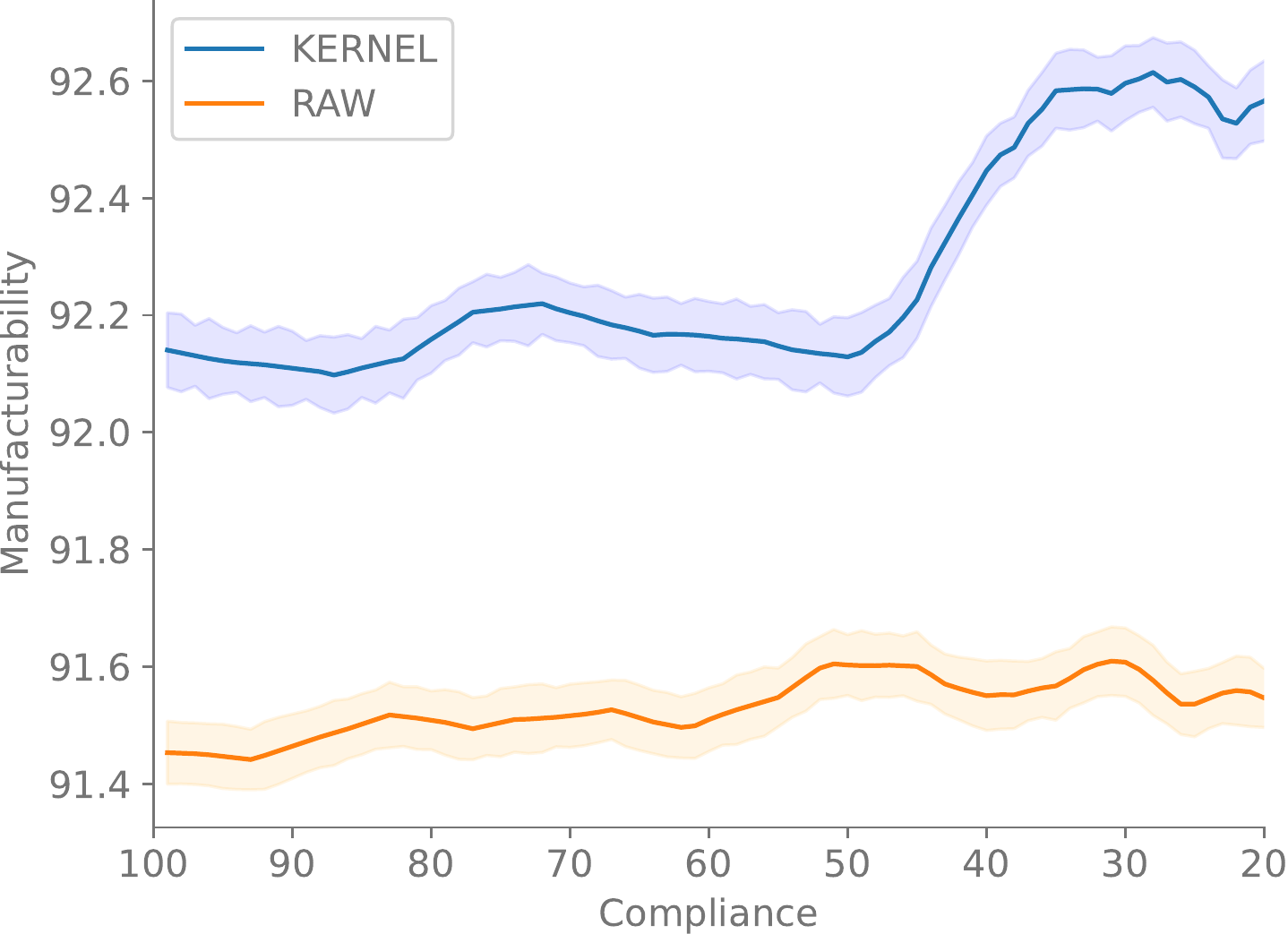}
    \caption{Manufacturability as measured in terms of absence of floating material and load disrespect, and compliance lower than 100 for out-of-distribution constraints.}
    \label{fig:manufacturability}
\end{figure}
We define a topology as manufacturable if it satisfies all constraints, has no floating material, and compliance is below 100. 
In Figure~\ref{fig:manufacturability}, we depict how manufacturability, measured as the percentage of configurations that meet these criteria, varies with performance. We evaluate out-of-distribution constraints and compare models trained on raw loads and boundary conditions with those trained on kernel relaxation. Our results show that manufacturability improves with performance, i.e., lower average compliance, of generated topologies. Notably, we observed a higher fraction of manufacturable topologies when conditioning on the kernel relaxation, supporting our hypothesis that the kernel is an effective approximation for loads and boundaries.

\paragraph{Kernel Ablation.}

The table presents the results of a kernel ablation study for the TopoDiff-FF model. We train the models using short runs. Four different inverse distance kernels with different exponents are evaluated, and their performance is measured in terms of compliance error (CE), volume fraction error (VFE), floating material (FM), load disrespect (LD), and the percentage of compliant samples with a compliance value smaller than 100 ($conf|{c<100}$). Lower values for CE, VFE, FM, and LD indicate better performance, while higher values for $conf|{c<100}$ indicate better compliance with the given constraints. The results show that the $1/r^2$ kernel performs well overall, with the lowest CE and VFE values and the highest $conf|{c<100}$ percentage when learning the power using $\beta$. The $1/r^{\beta}$ kernel also performs well, with the lowest FM value, while the $1/r$ kernel performs the worst in all metrics.

\begin{table}[ht]
    \centering
    \setlength{\tabcolsep}{3pt}
    \begin{tabular}{c|c c c c c}
                     & CE $\downarrow$ & VFE $\downarrow$ & FM $\downarrow$ & LD $\downarrow$ & $conf|_{c<t}$ $\uparrow$         \\
    \midrule
       $K(1/r)$         & 189.76 & 15.54 & 8.60 & 1.4 & 48.80 \\
       $K(1/r^2)$       & 47.62 & 2.38 & 9.23  & 0.0 & 94.32  \\
       $K(1/r^4)$       & 55.01 & 2.35 & 8.15 & 0.0  & 95.90   \\
       $K(1/r^{\beta})$ &  36.20 & 2.22 & 8.63 & 0.0 & 95.66  \\
       \bottomrule
    \end{tabular}
    \caption{Kernel Ablation for TopoDiff-FF. All metrics in $\%$. We train models for a shorter time on a subset of the training data.}
    \label{tab:kernel-ablation}
\end{table}

\section{Discussion} 
It is essential to acknowledge the limitations of purely data-driven approaches, such as TopoDiff-FF, which struggle to generalize to out-of-distribution scenarios since these models lack a way to generalize to new situations not encountered during training. In contrast, TopoDiff incorporates additional conditioning information obtained from numerical analysis and the underlying physics of the problem, which enables the model to generalize to regions of the domain not included in the training data.
The kernel proposed for TopoDiff-FF is versatile and can be easily applied to any domain, resolution, and material, making it a valuable tool for engineering design. Contrary, FEM conditioning requires a thorough understanding of the problem, and it does not scale well with resolution and problem complexity. Therefore, TopoDiff-FF represents a powerful alternative for engineering design, particularly in situations where FEM conditioning is not feasible or effective.
Furthermore, our results demonstrate that combining the topology generated by TopoDiff-FF with a few iterations of SIMP can produce better results than either data-driven or physics-informed methods, indicating a way to increase expressivity for such models in engineering design.

\paragraph{Limitations.}
Deep Generative models for topology optimization are a promising direction to improve efficiency, scalability, and variety in engineering design but they still present limitations ~\citep{woldseth2022use}.
In general, three main objectives can be identified for data-driven approaches: direct design, speedup, and upsampling. 
The authors in~\cite{woldseth2022use} are critical of end-to-end approaches for direct design, stating that most of the methods proposed in the literature produce poor designs and are expensive and limited in the variety of problems and mesh resolutions they can handle. This is indeed a good critique because most methods focus on a specific domain and resolution. 
The authors argue that the iterative nature of topology optimization is not what makes it impractical but rather the computational load of the costly components within each iteration. They suggest that the focus should shift towards alleviating this computational load. 
The authors also suggest that the main contribution of the learning methods should be in speeding up the computations in the intermediate steps of the iterative optimization process or post-processing optimized results for manufacturability.
Where most of these critiques are accurate, there is a new wave of generative models that are tackling such challenges and solving or alleviating most of these issues, from sampling efficiency to reliance on a specific domain; to optimization-based refinement and direct design with performance awareness~\cite{regenwetter2022deep}.

\section{Conclusion}
In summary, our method of utilizing kernel approximation as conditioning for generative models and refining with optimization techniques has shown promise in enhancing precision, constraint satisfaction, manufacturability, and performance. To further improve our approach, future research could explore methods for maintaining diversity while preserving precision and constraint satisfaction. Furthermore, the scalability of our approach to larger and more complex problems should be investigated. In conclusion, we believe that our work offers new possibilities for advancing generative design in topology optimization and engineering design.

\bibliographystyle{_asmeconf/asmeconf}
\small
\bibliography{biblio.bib}

\onecolumn
\appendix
\section{Examples of Generated Topologies}

\begin{figure*}[ht]
    \centering
    \includegraphics[width=.9\linewidth, height=18cm]{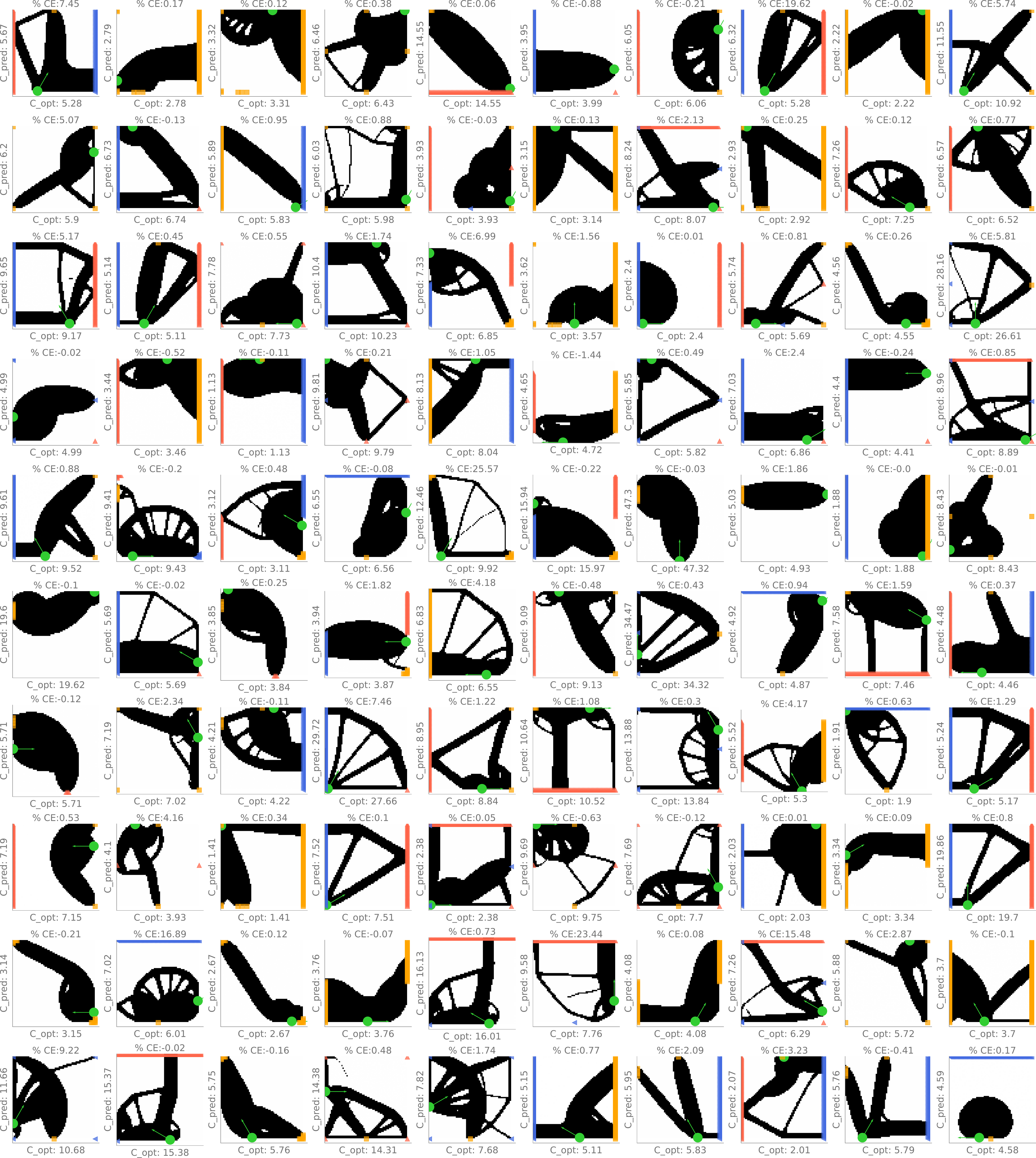}
    \caption{Examples of generated topologies with good performance.}
    \label{fig:low-compliance}
\end{figure*}

\end{document}